\documentclass[journal]{IEEEtran}
\newcommand{\thickhline}{%
    \noalign {\ifnum 0=`}\fi \hrule height 1pt
    \futurelet \reserved@a \@xhline
}

\usepackage{graphicx}
\usepackage{enumitem}
\usepackage{siunitx}
\usepackage{multirow}
\usepackage{xcolor}
\usepackage[normalem]{ulem}
\usepackage{color}

\usepackage[ruled]{algorithm2e} 

\usepackage{booktabs} 

\usepackage{CJK}
\usepackage{amsthm,amsmath,amssymb}
\usepackage{url}

\ifCLASSOPTIONcompsoc
  \usepackage[nocompress]{cite}
\else
  \usepackage{cite}
\fi
\ifCLASSINFOpdf
\else
\fi
\begin{document}
%
\title{Fine-Grained Vehicle Perception via 3D Part-Guided Visual Data Augmentation}

\author{Feixiang Lu, Zongdai Liu, Hui Miao, Peng Wang, Liangjun Zhang, Ruigang Yang,~\IEEEmembership{Senior Member,~IEEE,} \\ Dinesh Manocha,~\IEEEmembership{Fellow,~IEEE,} and Bin Zhou
}

%
%

\markboth{Journal of \LaTeX\ Class Files,~Vol.~14, No.~8, July~2020}%
{Shell \MakeLowercase{\textit{et al.}}: Bare Demo of IEEEtran.cls for Computer Society Journals}
%



\IEEEtitleabstractindextext{%
\begin{abstract}
Holistically understanding an object and its 3D movable parts through visual perception models is essential for enabling an autonomous agent to interact with the world. For autonomous driving, the dynamics and states of vehicle parts such as doors, the trunk, and the bonnet can provide meaningful semantic information and interaction states, which are essential to ensuring the safety of the self-driving vehicle. Existing visual perception models mainly focus on coarse parsing such as object bounding box detection or pose estimation and rarely tackle these situations. In this paper, we address this important autonomous driving problem by solving three critical issues. First, to deal with data scarcity, we propose an effective training data generation process by fitting a 3D car model with dynamic parts to vehicles in real images before reconstructing human-vehicle interaction (VHI) scenarios. Our approach is fully automatic without any human interaction, which can generate a large number of vehicles in uncommon states (VUS) for training deep neural networks (DNNs). Second, to perform fine-grained vehicle perception, we present a multi-task network for VUS parsing and a multi-stream network for VHI parsing. Third, to quantitatively evaluate the effectiveness of our data augmentation approach, we build the first VUS dataset in real traffic scenarios (\textit{e.g.}, getting on/out or placing/removing luggage). Experimental results show that our approach advances other baseline methods in 2D detection and instance segmentation by a big margin (over 8\%). In addition, our network yields large improvements in discovering and understanding these uncommon cases. Moreover, we have released the source code, the dataset, and the trained model on Github (\url{https://github.com/zongdai/EditingForDNN}).

\end{abstract}

}

\maketitle

\IEEEdisplaynontitleabstractindextext

%
\IEEEpeerreviewmaketitle

\section{Introduction}
\label{sec:intro}
\begin{figure}
\centering
\includegraphics[width=0.95\linewidth]{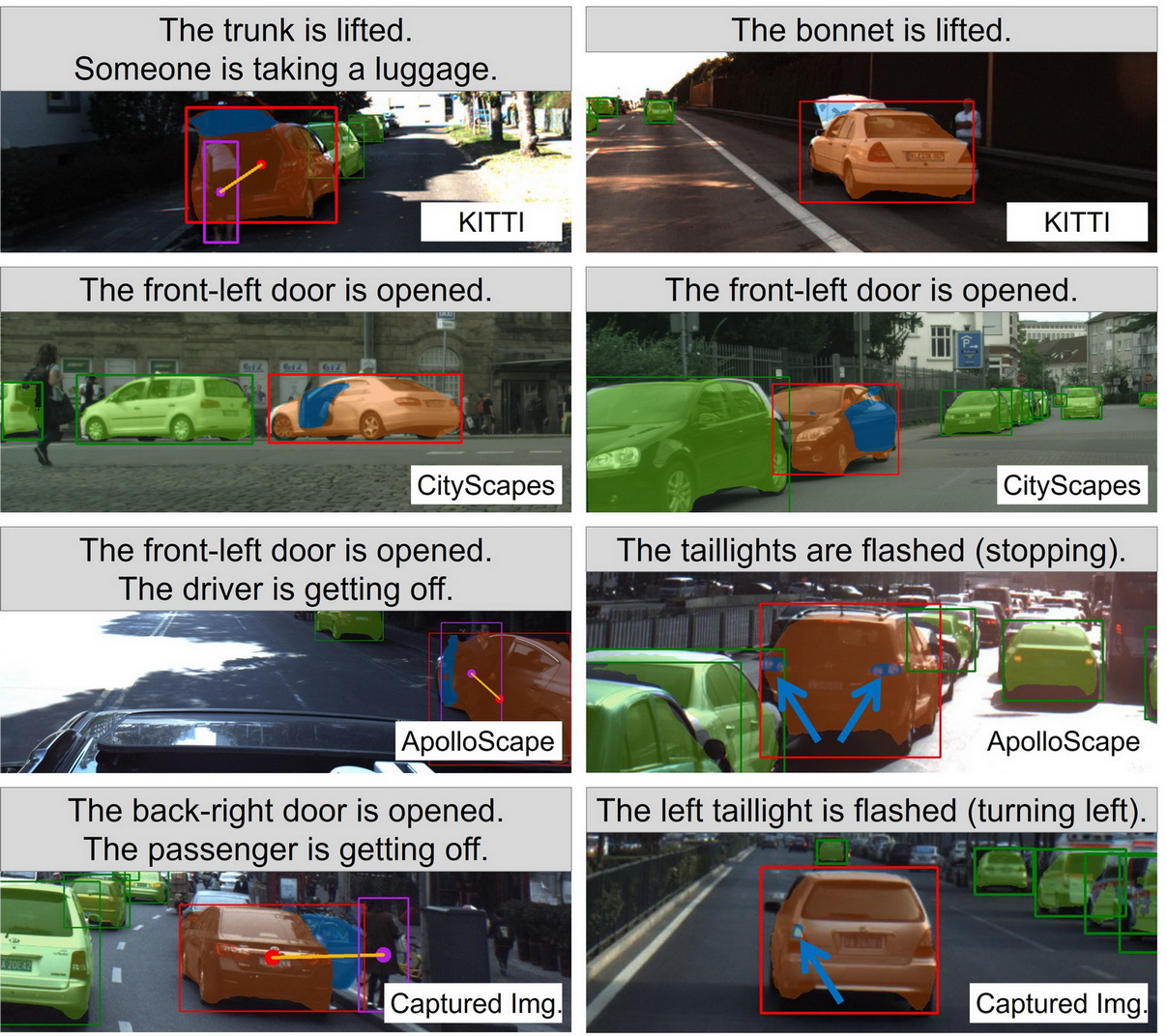}
\caption{Fine-grained vehicle perception on various AD-oriented datasets. Our perceiving results include 2D detection (red bounding box), instance segmentation (orange mask), dynamic part segmentation (blue mask), and state description. In addition, we detect the human (pink bounding box) as he or she interacts with the vehicle (yellow line). Note that the common-state vehicles are in green.}
\label{Fig:Teaser}
\end{figure}

\IEEEPARstart{A}UTONOMOUS driving (AD) has long been considered one of the most exciting technologies that artificial intelligence (AI) is expected to deliver. Parsing and analyzing moving objects in particular vehicles is an important problem in the context of AD systems. In contrast to other generic objects, vehicles are composed of articulated and movable 3D parts (\textit{e.g.}, doors, the trunk, and the bonnet). The dynamics and states of these parts can potentially provide meaningful semantic information and interactive vehicle states, which are essential for ensuring the safety of the self-driving vehicle. For example, when a car parked on the shoulder of the road has its door opened, it is likely that someone will step out of the car. As a response, the autonomous vehicle should perform a proper action such as slowing down or changing lanes to avoid hitting the individual. While such cases are not common, they can be dangerous or deadly if there is no understanding of such scenarios in the AD systems. As illustrated in Fig.~\ref{Fig:Teaser}, there are many such cases in real driving scenarios, and these are extremely challenging for existing AD systems to perceive and handle.

In this paper, we mainly deal with the problem of \textit{fine-grained} vehicle perception from a single image, particularly focusing on \textit{vehicles in uncommon states} (\textit{VUS}) in AD scenarios, by providing a 2D perception model that enables detailed 3D part-level parsing and understanding for vehicles and their interaction states with humans. Specifically, the scope of ``fine-grained vehicle perception'' includes: 1) recognition of vehicles in common and uncommon states, 2) 2D detection, 3) instance-level segmentation, 4) dynamic part segmentation, 5) state description, and 6) interaction semantics between vehicles and humans. Figs.~\ref{Fig:Teaser}, \ref{fig:sequence}, \ref{fig:site}, \ref{fig:city} show the results of fine-grained perception. Prior works in this area have studied one or two tasks such as 2D detection (\textit{e.g.}, \cite{ren2015faster, fu2017dssd}), semantic/instance-level segmentation (\textit{e.g.}, \cite{he2017mask}) and part level segmentation (\textit{e.g.}, \cite{lu2018beyond}). Although these state-of-the-art (SOTA) methods have achieved good results for vehicle perception, there are many challenges with respect to uncommon vehicle identification, dynamic part segmentation, and \textit{``vehicle-human'' interaction (VHI)} semantics. In summary, the limitations of these prior works include the following: 1) they are unable to recognize the VUS cases; 2) the results of detection and segmentation are not accurate due to the dynamic/movable parts; 3) it is difficult to parse the VHI semantics which can provide more information for planning and decision modules of AD systems; and 4) there is high coupling between the training data and network. In other words, it is difficult to train a network to perform 6 perception tasks using a uniform dataset.

To achieve fine-grained vehicle perception, we evaluate many popular AD datasets, including KITTI~\cite{geiger2013vision}, CityScapes~\cite{cordts2016cityscapes}, ApolloScape~\cite{huang2018apolloscape}, and ApolloCar3D~\cite{song2019apollocar3d}. As shown in Fig.~\ref{Fig:Teaser}, we first find cases where a piece of car has moved in real driving scenarios. Second, we determine that the number of cases is too scarce to train an effective model for fine-grained 3D part understanding, \textit{i.e.} only tens within 15,000 images in KITTI. The most common strategy for generating enough training data is manually crowd sourcing large numbers of real images~\cite{geirhos2018imagenet}, which is labor-intensive. However, other solutions such as obtaining a dataset with a simulated environment and computer graphics~\cite{alhaijaaugmented, tremblay2018training, prakash2018structured} will produce strong domain gaps between vehicle and scene appearances and realistic scenarios.

\subsection*{Main Contributions}
We present a novel approach for fine-grained vehicle perception from a single image, with particular respect to \textit{vehicles in uncommon states} (\textit{VUS}). This includes a new method for data augmentation and two efficient networks for vehicle parsing. The key aspect of our method is the 3D part-guided data augmentation strategy, which first fits a 3D vehicle geometric model with dynamic parts in images and then re-renders the edited vehicle with re-configured parts and a realistic texture. Specifically, our 3D models are from ApolloCar3D~\cite{song2019apollocar3d}, which provides the annotations of the CAD model and the 6-DoF pose for each 2D vehicle instance. Then we segment out 4 semantic parts (\textit{i.e.} two headlights and two taillights) and 6 movable parts (\textit{i.e.} four doors, the bonnet, and the trunk) for each 3D model. Furthermore, for each movable part, we annotate its motion axis and constrain its range of movement. We use these 3D parts as a guide to directly edit the 2D vehicles from source images, which can automatically generate a lot of VUS samples for the training network.

We further observe that humans always interact with the vehicle parts (\textit{e.g.}, getting out of the car, taking luggage out of the open trunk, etc.). To enhance the fidelity of the synthetic data, we further generate \textit{vehicle-human interaction (VHI)} data. Specifically, we first reconstruct the human 3D motion sequence by fitting the SMPL \cite{loper2015smpl} model to the captured RGB video using the VIBE approach \cite{kocabas2020vibe}. Then we generate a large set of texture maps based on the human database. Finally, we render the SMPL model with different texture maps to integrate the vehicle parts, yielding a large number of vehicle-human interaction samples. Experiments show that the generated VHI data can effectively improve the performance of vehicle perception.

Based on the augmented data, we present a new multi-task network to perform fine-grained vehicle perception. Our network can simultaneously output the results of 2D detection, instance segmentation, dynamic part segmentation, and state description.
Our deep model is significantly better at understanding vehicles in AD than models without our generated dataset. Moreover, we design a multi-stream fusion network to parse the VHI semantics, including getting on/out, placing/removing luggage, etc.

Finally, to benchmark our perception model, to the best of our knowledge, we have built the first VUS dataset with a large number of described uncommon states of vehicles, which annotates 1850 VUS instances from 1441 real-traffic images. We conduct various quantitative and qualitative experiments to justify the effectiveness of our approach for fine-grained vehicle perception.

In summary, our contributions include:
\begin{itemize}

\item We present a 3D part-guided visual data augmentation pipeline for automatic training data generation, including VUS data and VHI data, which helps to learn fine-grained vehicle perception models in AD.

\item We propose a multi-task network that outputs both instance-level and part-level vehicle understanding. Moreover, we implement a multi-stream fusion network to parse the vehicle-human interaction semantics. These fine-grained perception results are important to the safety of autonomous vehicles.

\item To benchmark our visual data augmentation approach, we have constructed the first VUS dataset with 1441 real images with fine-grained annotation of vehicles and humans in many uncommon states to demonstrate the effectiveness of our approaches.
\end{itemize}

\begin{figure*}
\begin{center}
   \includegraphics[width=0.95\linewidth]{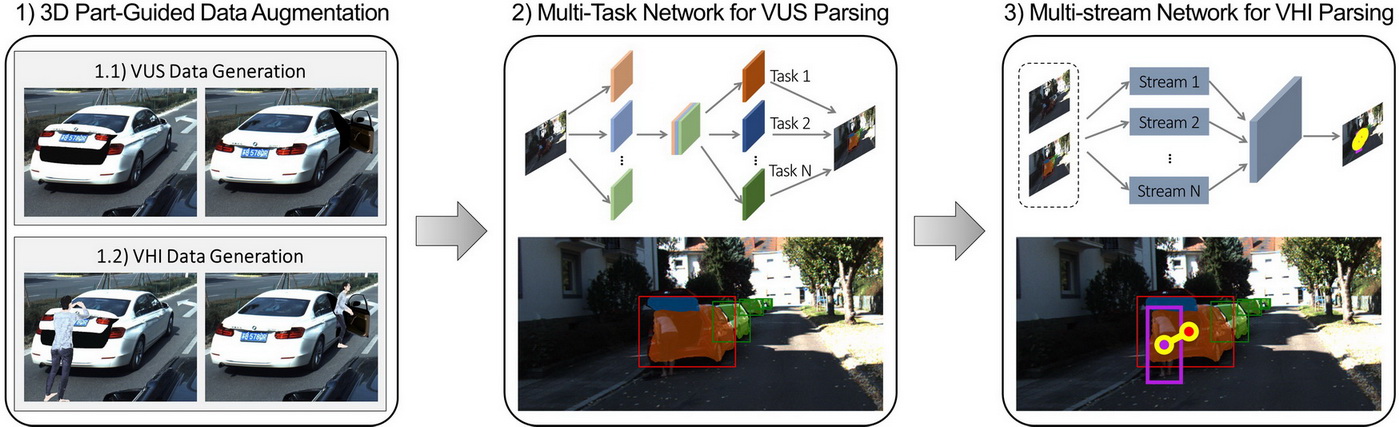}
\end{center}
   \caption{Our visual data augmentation pipeline includes three main components: 1) 3D part-guided data augmentation, 2) a multi-task network for VUS parsing, and 3) a multi-stream network for VHI parsing. This pipeline provides a complete and fully automatic solution for fine-grained vehicle perception, which can not only generate the visual data for training, but also present end-to-end deep networks for VUS parsing and VHI parsing.  }
\label{fig:pipeline}
\end{figure*}

An earlier version of this work was published in CVPR 2020 \cite{liu20203d}. In the current paper, we propose a visual data augmentation pipeline that is a complete and fully automatic solution for fine-grained vehicle perception (shown in Fig.~\ref{fig:pipeline}). In particular, we further generate the vehicle-human interaction (VHI) data for vehicle parsing. By using both VUS and VHI data, we train a multi-stream fusion network to parse vehicle-human interaction semantics. Experiments demonstrate that our data augmentation pipeline is robust and effective. In addition, the generated VHI data can effectively improve the performance of fine-grained vehicle perception (\textit{i.e.} the CVPR version) on 2D detection, instance-level segmentation, part-level segmentation, and state description by 3.2, 1.8, 3.1, and 1.9 percentage points, respectively.  To facilitate further research and reproduce our work easily, we have released the source code, datasets, and the trained model on Github: \url{https://github.com/zongdai/EditingForDNN}.

The remainder of this paper is structured as follows. Sec.~\ref{sec:related_work} reviews related work. Sec.~\ref{sec:pipeline} introduces our data augmentation pipeline for fine-grained vehicle perception. We describe the constructed VUS dataset in Sec.~\ref{sec:vus_dataset}. We compare our approach with state-of-the-art methods on VUS datasets and discuss the applications to AD in Sec.~\ref{sec:results}.

\section{Related Work} 
\label{sec:related_work}
Fine-grained vehicle perception is important in ensuring the safety of autonomous vehicles, and it can provide useful information for planning and decision modules of AD systems. We review the related works in the following section.

\subsection{Datasets for Autonomous Driving}
Several datasets have been constructed and released that focus on perception in AD. These datasets can be divided into two categories: real datasets and synthetic datasets. For real datasets (\textit{e.g.}, CamVid~\cite{brostow2009semantic}, KITTI~\cite{geiger2013vision}, CityScapes~\cite{cordts2016cityscapes}, Toronto~\cite{wang2016torontocity}, Mapillary~\cite{neuhold2017mapillary}, and BDD100K~\cite{yu2018bdd100k}, ApolloScape~\cite{huang2018apolloscape}, and ApolloCar3D~\cite{song2019apollocar3d}), the source data are captured from real traffic scenarios. For the synthetic datasets (\textit{e.g.}, SYNTHIA~\cite{ros2016synthia}, P.F.B.~\cite{richter2017playing}, and Virtual KITTI~\cite{engelmann2017exploring}), the data are generated by 3D rendering using computer graphics techniques. However, the annotations of these datasets are only focused on common vehicles, which ignore the VUS cases.

\subsection{Data Generation for Deep Network}
The effectiveness of the trained deep models largely relies on the network architecture (\textit{e.g.}, AlexNet~\cite{krizhevsky2012imagenet}, VGG~\cite{simonyan2014very}, and ResNet~\cite{he2016deep}) and the training data. However, it is not easy to build real datasets~\cite{deng2009imagenet,lin2014microsoft,huang2018apolloscape},   and the process is laborious, costly, and inefficient. Therefore, researchers use the synthetic data to train deep networks. 3D-based methods directly render 3D models with pre-defined textures and illuminations to generate the 2D images with ground-truth annotations to learn deep networks. However, pre-building complex and diverse 3D scenes is time-consuming. Therefore, 2D-based methods directly cut the foreground objects and paste them to other backgrounds (\textit{e.g.}, \cite{dwibedi2017cut}). However, these ``cut-paste'' methods are limited in terms of occlusion and shadow.

Nevertheless, synthetic data improves many computer vision tasks such as optical flow~\cite{butler2012naturalistic, dosovitskiy2015flownet}, scene flow~\cite{mayer2016large}, stereo~\cite{qiu2016unrealcv, zhang2016unrealstereo}, semantic segmentation~\cite{richter2016playing, ros2016synthia}, 3D keypoint extraction~\cite{suwajanakorn2018discovery}, viewpoint~\cite{su2015render}, object pose~\cite{muller2018sim4cv}, 3D reconstruction~\cite{handa2016understanding}, and object detection~\cite{alhaijaaugmented, gaidon2016virtual, prakash2018structured, tremblay2018training}.  
The key problem for these works is how to decrease the domain gap between the synthetic data and the real data. For example, researchers use domain adaptation/randomization approaches (\textit{e.g.}, \cite{tobin2017domain}) to get the optimal results for the task of vehicle detection~\cite{prakash2018structured, tremblay2018training}. Alhaija \textit{et al.}~\cite{alhaijaaugmented} combine the strengths of 3D- and 2D-based approaches to augment the rendering vehicles to the real traffic backgrounds to generate photo-realistic images. Hinterstoisser \textit{et al.}~\cite{Hinterstoisser_2018_ECCV_Workshops} show that only using synthetic data, they can train an effective detector by freezing a pre-trained feature extractor.

\subsection{Fine-Grained Vehicle Parsing}
To ensure the safety of autonomous vehicles, it is essential that they are able to perceive the surroundings, particularly the moving vehicles, with fine granularity. Existing methods can be divided into four main classes: 1) 2D bounding boxes detection (\textit{e.g.}, SSD513~\cite{fu2017dssd}, YOLOv3~\cite{redmon2018yolov3}, Faster-RCNN~\cite{ren2015faster}); 2) semantic/instance segmentation (\textit{e.g.}, Mask-RCNN~\cite{he2017mask}); 3) keypoints regression (\textit{e.g.}, ApolloCar3D~\cite{song2019apollocar3d}); and 4) part segmentation (\textit{e.g.}, \cite{wang2015joint, xia2016zoom, lu2018beyond}). These prior works can get good results in individual tasks. However, they are limited in performing fine-grained vehicle perception, particularly in terms of parsing and understanding the vehicles in uncommon states.

\begin{figure*}
\begin{center}
   \includegraphics[width=0.98\linewidth]{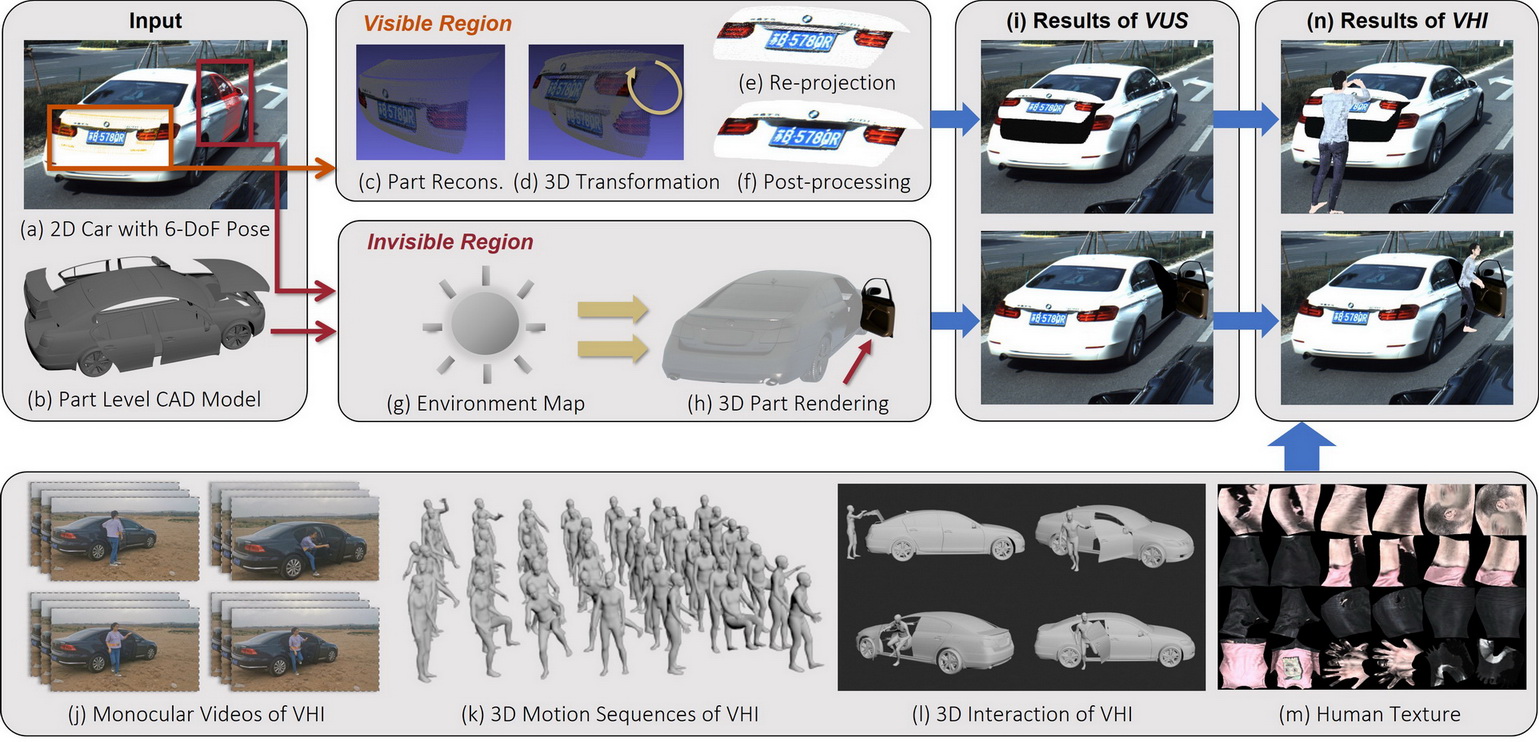}
\end{center}
   \caption{This is an overview of 3D part-guided data augmentation, including VUS and VHI data generation. For the visible region of VUS data, we take the posed 2D car instances (a) and part-level CAD models (b) as input. Then we reconstruct the 3D part (c) to rotate and translate it into a new state (d). The transformed 3D part is projected onto the image space, and then a post-processing method is used to refine the 2D results. For the invisible region, we use the environment map (g) to render the reverse side of the 3D part (h). (f) and (h) are augmented to the source image to generate the results of VUS (i). For VHI data augmentation, we first capture a lot of monocular videos of VHI (j), which are leveraged to estimate the 3D motion sequences (k). Next, we use the motion sequence to set the human body to virtually interact with the vehicle in 3D space (l). In addition, we build a human texture dataset (m). We randomly select the texture map to render the 3D human body of (l) to obtain the final results of VHI (n). }
\label{fig:overview}
\end{figure*}

\section{Visual Data Augmentation Pipeline}
\label{sec:pipeline}
In this section, we present the visual data augmentation pipeline, which includes a data generation approach and two networks for data learning (shown in Fig.~\ref{fig:pipeline}). Specifically, we first introduce the 3D part-guided data augmentation approach, which can automatically generate the VUS data and the VHI data for training (Sec.~\ref{sec:data_augmentation}). Then we propose a novel multi-task network for fine-grained vehicle perception (Sec.~\ref{sec:vus_parsing}). To obtain the vehicle-human interaction semantics, we further design a multi-stream network for VHI parsing (Sec.~\ref{sec:vhi_parsing}).

\subsection{3D Part-Guided Data Augmentation}
\label{sec:data_augmentation}
As demonstrated in Fig.~\ref{fig:overview}, we leverage the 3D parts to generate the VUS data and the VHI data. While most components of our data augmentation pipeline are automatic, the manual work is related to the data pre-processing, including: 1) CAD model segmentation and motion axis annotation (Sec.~\ref{sec:data_augmentation_VUS}) and 2) video sequencing of VHI (Sec.~\ref{sec:data_augmentation_VHI}). When these data are ready, our approach is fully automatic without any human interactions from input (source images) to output (the VUS data and the VHI data).

\subsubsection{Synthesizing VUS Data}
\label{sec:data_augmentation_VUS}

\begin{figure}
\begin{center}
   \includegraphics[width=0.98\linewidth]{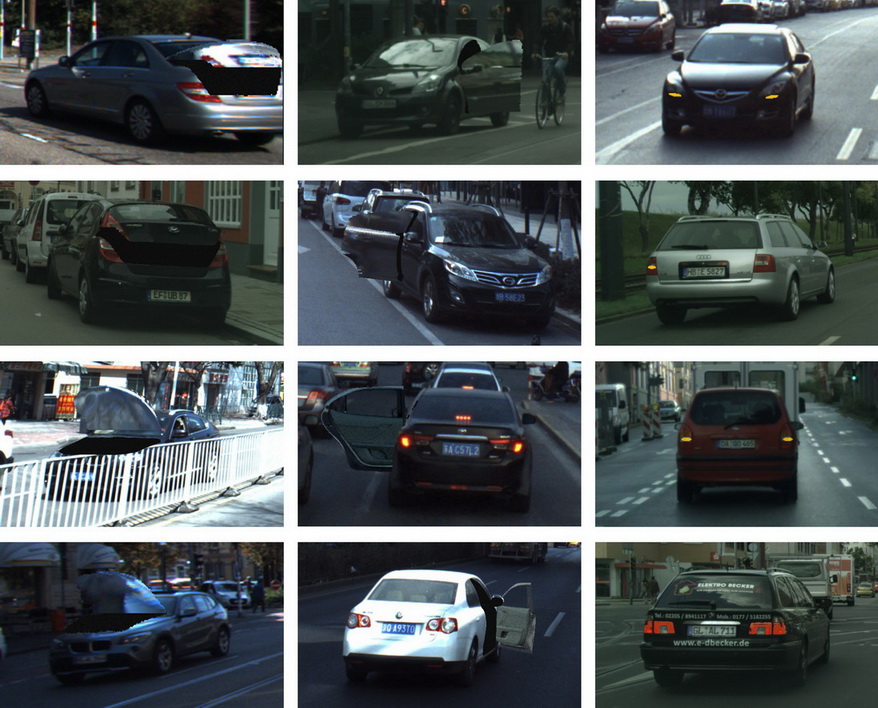}
\end{center}
   \caption{The VUS data generated by our approach. The first column and the second column show the editing results of movable parts (\textit{i.e.} four doors, the trunk, and the bonnet). The third column shows the editing results of the semantic parts (\textit{i.e.} two headlights and two taillights).}
\label{fig:editing_results_VUS}
\end{figure}

We use the 3D vehicle parts to automatically edit the source 2D images to generate various VUS samples (Fig.~\ref{fig:overview} (a)$\sim$(h)). The input is the 2D/3D aligned vehicle data with annotated CAD models and 6-DoF poses from ApolloCar3D~\cite{song2019apollocar3d}. Then we segment those 3D models to obtain the \textit{movable parts} (\textit{i.e.} bonnet, trunk, and four doors) and the \textit{semantic parts} (\textit{i.e.} two headlights and two taillights). For the semantic parts, we directly project them to the image space to obtain the corresponding 2D regions, which are further edited to yellow or red flashing effects (the third column in Fig.~\ref{fig:editing_results_VUS}). For the movable parts, we first annotate their motion axis, then transform the 3D parts to guide 2D image editing.

Note that these CAD models provided by ApolloCar3D are low-quality. It is difficult to unfold their texture maps to perform 3D rendering. Instead, we generate the synthesized results using the image editing techniques. Specifically, we first render the 3D movable parts to generate a corresponding depth map $D$, according to the global rotation $\mathbf{R_g}$, translation $\mathbf{t_g}$, and the camera intrinsic matrix  $\textbf{K}$. For each 2D pixel $\mathbf{u} = (u,v)^\top$ with depth value $D(\mathbf{u})$, we  convert it to acquire 3D point $\mathbf{P} = (x,y,z)^\top$ through
\begin{equation}
\mathbf{P} = \mathbf{R_g^{-1}} \cdot \big(D(\mathbf{u}) \cdot \mathbf{K^{-1}} \cdot \mathbf{\dot{u}} - \mathbf{t_g} \big).
\end{equation}
Here, $\mathbf{\dot{u}}$ is a homogeneous vector: $\mathbf{\dot{u}} = (\mathbf{u}^\top|1)^{\top}$.

Assuming the part is locally transformed with a 3D rotation $\mathbf{R_o}$ along with the motion axis, the axis translates $\mathbf{t_o}$ in the global coordinate. We compute the pixel's new position $\mathbf{u'}$ in the image domain, which is defined as:
\begin{equation}
\mathbf{u'} = \bigg \lfloor \pi \Big( \mathbf{K} \cdot \big(\mathbf{R_g}(\mathbf{R_o}(\mathbf{P}-\mathbf{t_o})+\mathbf{t_o} \big)+\mathbf{t_g} \big) \Big) \bigg\rfloor.
\end{equation}
Here, the function $\mathbf{u} = \pi(\mathbf{P})$ performs perspective projection of $\mathbf{P}\in \mathbb{R}^3 = (x,y,z)^\top$ including dehomogenization, to obtain $\mathbf{u}\in \mathbb{R}^2 = (x/z, y/z)^\top$.

Note that the transformed pixels are always sparse and noisy in the part region (Fig.~\ref{fig:overview} (e)). Here, we call the non-valued pixel a ``\textit{hole.}'' In order to fill these holes, we perform the linear blending algorithm~\cite{sumner2007embedded} to obtain the RGB values.
In general, pixels that are close to one another will be the most similar. Thus, for consistency and efficiency, we limit the influence of the valued pixels on a particular hole to the K-nearest neighbors. The weights $\omega_i$ for each pixel $\mathbf{u}_i$ are pre-calculated as
 \begin{equation}
 \omega_i(\mathbf{h}) = \Big(1 -  \left \| \mathbf{h} - \mathbf{u}_i  \right \|_2^2 / d_{max}\Big)^2
 \end{equation}  
 and then normalized to sum to one. Here, $d_{max}$ is the distance to the $(k+1)$-nearest pixel. We set $K=8$ in our experiments. The RGB value of the hole $h$ is calculated according to 
 \begin{equation}
 C(\textbf{h}) = \sum_{i=1}^K \omega_i(\mathbf{h}) \cdot C(\mathbf{u}_i).
 \end{equation}
After interpolating the non-valued pixels, we apply a bilateral filter~\cite{tomasi1998bilateral} on the synthetic images. The smoothed results are shown in Fig.~\ref{fig:overview} (f) and Fig.~\ref{fig:editing_results_VUS}.

\subsubsection*{Generating Invisible Vehicle Region}
For the case of an opening car door, we can generate visually compelling results if the car is facing the camera. When the car is facing in the opposite direction, however, an opening door will introduce some invisible regions in the original image. We can divide these invisible regions into two classes: one is the reverse side of the part and the other is the vehicle interior (\textit{e.g.}, seat and steering wheel). Empirically, the interior regions are always dark due to inadequate illumination. Therefore, we directly fill interior regions in with gray. While we have also attempted to use random colors and patches from real images according to the experimental results, we do not find obvious differences between the options.

Compared with coloring the interior regions, coloring the reverse side of the part is rather complex. It is not appropriate to directly fill the region in with pure color. Thus, we adopt the photo-realistic rendering pipeline to generate high-fidelity results of the reverse side. We first construct a small, expert-designed 3D model database for movable parts. The part materials are manually labeled and the BRDF parameters are pre-defined by a professional artist. Then we compute the environment map \cite{lalonde2010synthesizing} of ApolloCar3D to perform photo-realistic rendering (Fig.~\ref{fig:overview} (i)). More synthetic results of VUS are shown in Fig.~\ref{fig:editing_results_VUS}.

\begin{figure}
\begin{center}
   \includegraphics[width=0.98\linewidth]{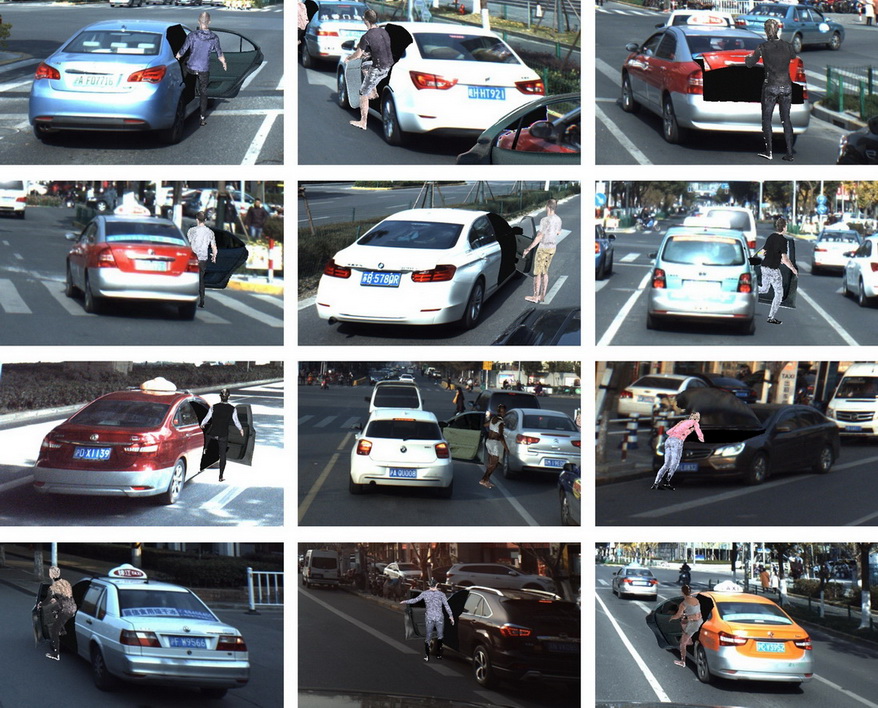}
\end{center}
   \caption{The VHI data generated by our approach. We highlight how we can synthesize various VHI cases, including getting in/out of the vehicle, placing/removing luggage, repairing the engine, and so on. These VHI cases are essential to ensuring the safety of autonomous vehicles. }
\label{fig:editing_results_VHI}
\end{figure}

\subsubsection{Synthesizing VHI Data}
\label{sec:data_augmentation_VHI}
In the real AD scenarios, vehicles in uncommon states are usually associated with human interactions (\textit{e.g.}, getting in/out of the car, placing/removing luggage, etc.). Perceiving such ``vehicle-human'' interaction (VHI) cases is essential to the safety of the autonomous vehicle. However, such VHI data is not easy to capture and collect while ensuring its diversity.  In this subsection, we introduce a method for automatically generating a large amount of VHI data for network training.

The VHI data augmentation pipeline is shown in Fig.~\ref{fig:overview}. We first use a hand-held camera to capture 100 monocular video sequences of VHI  (Fig.~\ref{fig:overview} (j)), such as getting in/out of the vehicle, placing/removing luggage, and opening the door/bonnet/trunk. Then a 3D parametric human template model (\textit{i.e.}, SMPL-X \cite{loper2015smpl}) is used to fit these videos through the VIBE approach \cite{kocabas2020vibe}, yielding the 3D human body sequences with motion parameters (Fig.~\ref{fig:overview} (k)). Next, the geometry mesh of SMPL-X is unfolded to obtain its corresponding UV map using the Unfold3D tool. We synthesize a lot of human texture maps using the SOTA approaches (\textit{i.e.} DensePose \cite{alp2018densepose} and SURREAL \cite{varol2017learning}) to construct the texture database.

For each edited vehicle part, we select an appropriate motion sequence to guide the SMPL-X model deformed with a ``coarse-to-fine'' pipeline. Specifically, we first rotate and translate the global human model to match the vehicle part. Then, we adjust the body pose (\textit{e.g.}, hand) to seamlessly interact with the vehicle parts (Fig.~\ref{fig:overview} (l)). After obtaining the posed SMPL-X model, we assign a random texture map using the constructed texture database (Fig.~\ref{fig:overview} (m)). Finally, we render the vehicle-human interaction models to generate the 2D images under the current view point (Fig.~\ref{fig:overview} (n), Fig.~\ref{fig:editing_results_VHI}). The proposed data augmentation approach for VHI is fully automatic, and is used to generate a large amount of training data for VUS parsing (Sec.~\ref{sec:vus_parsing}) and VHI understanding (Sec.~\ref{sec:vhi_parsing}).

\begin{figure}
\begin{center}
   \includegraphics[width=0.98\linewidth]{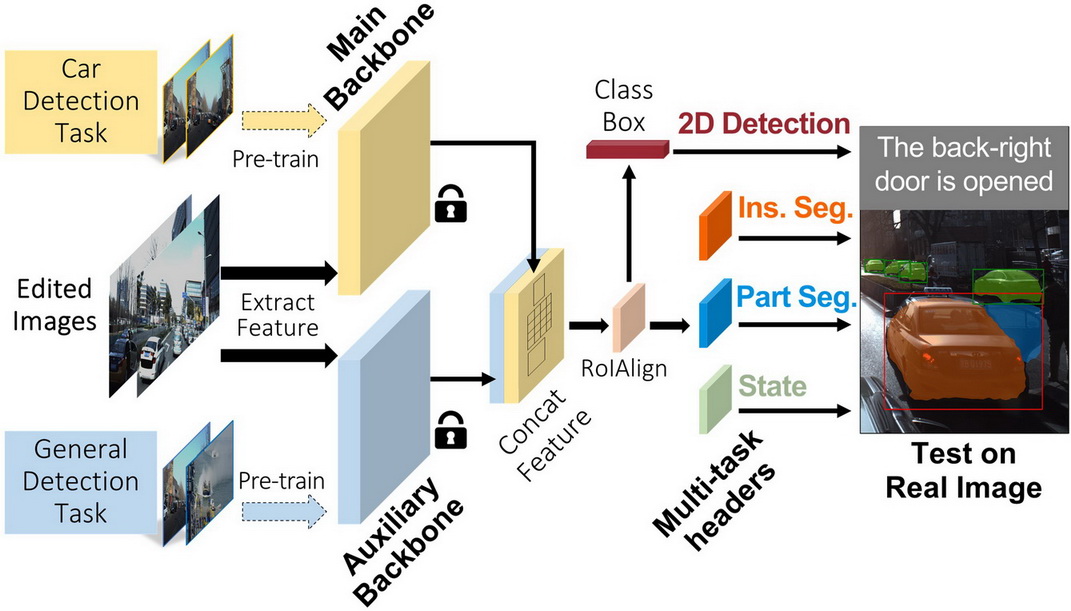}
\end{center}
   \caption{We design a two-backbone network to generalize training from synthetic to real VUS data. The main backbone is pre-trained on the ApolloCar3D dataset to extract the car body features. The auxiliary backbone is pre-trained on the COCO dataset to extract the general features of the edited region
(e.g. the rendered parts). We keep both backbones frozen and train our network,  which can output multiple results of 2D detection, instance-level  segmentation, dynamic part segmentation, and state description. In addition, we use the real-world images for testing to justify the effectiveness of our approach.}
\label{fig:network}
\end{figure}

\subsection{ Multi-Task Network for VUS Parsing }
\label{sec:vus_parsing}
Based on the synthetic data, we design a new multi-task network to perform VUS parsing. Our network can simultaneously output various results of 2D detection, instance segmentation, part segmentation, and state description (Fig.~\ref{fig:network}).

\subsubsection{Two Backbones}
Our goal is parsing VUS from real captured images by training only on the synthetic data (described in Sec.~\ref{sec:data_augmentation}). To achieve the transferability from synthetic data to real data, we use two backbones (\textit{i.e.} ResNet50-FPN~\cite{lin2017feature}) to extract features. The \textit{main backbone} is to extract the car body features, and it is pre-trained using Mask-RCNN~\cite{he2017mask} on the  ApolloCar3D~\cite{song2019apollocar3d} dataset and the CityScapes~\cite{cordts2016cityscapes} dataset guided by a car detection task. The \textit{auxiliary backbone} is used to extract the general features of the edited regions (\textit{e.g.}, the rendered parts), and it is pre-trained on the COCO dataset \cite{lin2014microsoft} guided by a general detection task. We fix the parameters of both backbones to train our network. In Sec.~\ref{sec::performance}, we conduct experiments to justify the effectiveness of this training strategy (\textit{i.e.} freezing two backbones).

\subsubsection{Dynamic Part Segmentation} 
We adopt Mask-RCNN~\cite{he2017mask} to implement the task of dynamic part segmentation. In Mask-RCNN, the mask branch outputs a $\mathit{Km^2}$ dimensional binary mask for each RoI aligned feature map, where $\mathit{K}$ is the number of classes and $\mathit{m}$ is the resolution. In addition, we take the dynamic part segmentation as a new channel, resulting in an output containing a $\mathit{(K+1)m^2}$ binary mask. Specifically, we feed a 14$\times$14 RoI aligned feature map to four sequential 256-d 3$\times$3 convolution layers. A 2$\times$2 deconvolution layer is used to up-sample the output to 28$\times$28. Finally, we define the $\mathit{L_{part}}$ as the average of per-pixel sigmoid cross-entropy loss.

\subsubsection{State Description}
We use a binary variable to represent the existence of the particular part state (\textit{i.e.} 1 for existence and 0 for others). Then, we define the ``part state vector'' as a concatenation of all the binary variables. Our method regresses the part state vector through the sequential convolution layers and a fully connected layer in the mask branch. Similarly, we define the $\mathit{L_{state}}$ as the average sigmoid cross-entropy loss.

\begin{figure}
\begin{center}
   \includegraphics[width=0.99\linewidth]{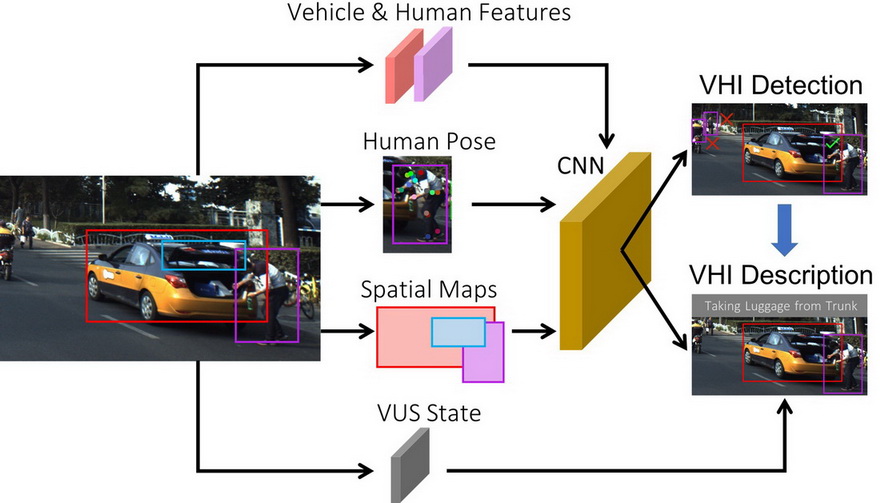}
\end{center}
   \caption{We design a multi-stream network to parse the vehicle-human interactions. We first utilize a human stream and a vehicle stream to extract human and vehicle features, respectively. Meanwhile, spatial maps and human pose maps are also calculated. We fuse these streams to a CNN network to detect the VHI cases. The VHI detection results are combined with the VUS states (Sec.~\ref{sec:vus_parsing}) to infer the interaction semantics between humans and vehicles.} 

\label{fig:hoinetwork}
\end{figure}

\begin{table*}[h]
	\caption{Components and details of the built VUS dataset, which annotates 1850 UVS instances from 1441 street-view images. ``fl-o.~(br-o.)'' indicates an opened front-left (back-right) part and ``l-tu.~(r-tu.)'' indicates turning left~(right). }  
  \begin{center}
	\begin{tabular}{c|c|c|c|c|c|c|c|c|c|c|c|c|c}
	\toprule[1pt]

	\multirow{2}{*}{\textbf{Datasets}} &  \textbf{Bonnet}  & \textbf{Trunk} &   \multicolumn{4}{c|}{ \textbf{Doors}}   &   \multicolumn{2}{c|}{ \textbf{Headlights}}   & \multicolumn{4}{c|}{ \textbf{Taillights}} & \multirow{2}{*}{\textbf{Total}}\\
	\cline{2-13}
	& lifted  & lifted & fl-o. &  fr-o. & bl-o. & br-o. &  l-tu. & r-tu.  &  l-tu. & r-tu.  & stop & alarm & \\
	\hline
	\textit{KITTI} & 1 & 9 & 1 & 0 & 0 & 5 & 1 & 0 & 2 & 1 & 8 & 0 & 28\\
	
	\textit{CityScapes} & 0 & 0 & 14 & 5 & 8 & 4 & 3 & 2 & 4 & 0 & 15 & 0 & 55\\
	
	\textit{ApolloScape} & 0 & 23 & 29 & 0 & 59 & 157 & 15 & 18 & 23 & 27 & 33 & 16 & 400\\
	
	\textit{ApolloCar3D} & 0 & 13 & 19 & 1 & 0 & 11 & 3 & 5 & 12 & 9 & 21 & 0 & 94\\
	
	\textit{Capt. Images} & 15 & 405 & 232 & 66 & 79 & 346 & 19 & 17 & 25 & 18 & 44 & 7 & 1273\\
	\hline
	\textbf{\textit{VUS Dataset}} & 16 & 450 & 295 & 72 & 146 & 523 & 41 & 42 & 66 & 55 & 121 & 23 & \textbf{1850}\\
	\bottomrule[1pt]
	\end{tabular}
	\label{tab::number_dataset}
  \end{center}
\end{table*}

\begin{figure*}
\begin{center}
   \includegraphics[width=1.0\linewidth]{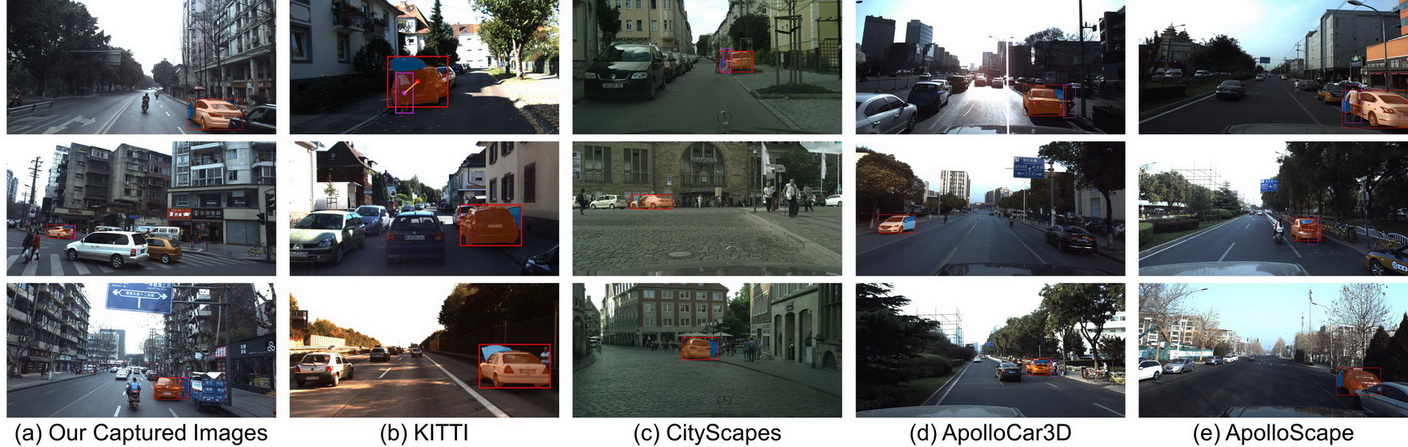}
\end{center}
   \caption{Samples of our constructed VUS dataset. For each 2D instance of VUS, we manually label its 2D bounding box (in red), instance segmentation (in orange), dynamic part segmentation (in blue), and state description. If there exist human interactions, we add the human bounding box (in pink), the connection between the center points of the human and the vehicle (with a yellow line), as well as the interaction semantics. We capture the majority of the source images (a). Only a few source images are from: (b) KITTI; (c) CityScapes; (d) ApolloCar3D; and (e) ApolloCar3D.}
\label{fig:dataset}
\end{figure*}

\subsubsection{Training Details}
First, we pre-train a Mask-RCNN with a ResNet50-FPN backbone both on ApolloCar3D~\cite{song2019apollocar3d} and CityScapes~\cite{cordts2016cityscapes}  through a car instance segmentation task. Then, we initialize the main backbone by copying the parameters of the pre-trained network. Simultaneously, we pre-train the auxiliary backbone on the COCO dataset using the same network architecture. Finally, we fix the parameters of these two backbones to train the network on the synthetic data. The multi-task loss is defined as:
\begin{equation}
\begin{split}
L = L^{p}_{class} + L^{p}_{reg} + L^{r}_{class} + L^{r}_{box}\\
+ L^{r}_{mask} + L^{r}_{state} + L^{r}_{part},
\end{split}
\end{equation}
where $\mathit{(.)^{p}}$ and $\mathit{(.)^r}$ indicate RPN and RCNN, respectively. The subscripts $\mathit{state}$ and $\mathit{part}$ denote the loss of state vector and part mask, respectively. We minimize our loss function using the SGD with a weight decay of 0.0001 and a momentum of 0.9. The learning rate is initially set to 0.002 and reduced by 0.1 for every 5 epochs.

\subsection{ Multi-Stream Network for VHI Parsing }
\label{sec:vhi_parsing}
The proposed multi-task network with two backbones in Sec.~\ref{sec:vus_parsing} is used to detect, segment, and parse the VUS. However, we notice that some VUS cases contain human interaction. Inevitably, if the vehicle is occluded by humans, it will degrade the network performance. To increase the safety of the AD and improve the perception performance, we present a multi-stream fusion network designed for inferring the vehicle-human interaction semantics, which is shown in Fig.~\ref{fig:hoinetwork}.  Specifically, we first utilize a human stream and a vehicle stream to extract human and vehicle features, respectively. Meanwhile, spatial maps\cite{chao2018learning} and human pose maps \cite{chao2018learning, li2019transferable} are also computed. We fuse these four streams to infer the interaction semantics between humans and vehicles.

\textbf{Human Stream}. We extract human appearance ROI pooling features using Mask-RCNN \cite{he2017mask}, then feed those features into the convolutional layer ($H_c$) to get the feature $f_h$. Then we employ pose estimation \cite{he2017mask} to estimate its 17 key-points (COCO format \cite{lin2014microsoft}). These key-points' 2D coordinates are reorganized into a 34-dimension vector that is rescaled to [0,1]. We exploit two 256 sized FCs to extract the pose feature $f_p$.

\textbf{Vehicle Stream}.  We further use the VUS parsing network (Sec.~\ref{sec:vus_parsing}) to extract the basic vehicle appearance ROI pooling features. In addition, we obtain high-level multi-task branch features. State description branch output is used to infer specific interaction patterns.

\textbf{Spatial Stream}.  The spatial stream is used to encode the spatial location of objects and humans. Candidate objects and human bounding boxes are encoded to a binary two-channel tensor, as in previous methods\cite{chao2018learning, li2019transferable}. Intuitively, such a human-object bounding box pair gives important cues for inferring interactive or non-interactive states, \textit{e.g.}, if a person is close to an object, it is very likely that they interact with each other. However, in urban street scenarios, there are many pedestrians around the vehicle, making it difficult to determine the real interactions. Therefore, we encode the vehicle dynamic part into an extra binary channel, which implies many more priors. These three channel tensors are fed into a convolutional layer ($S_c$) to get the feature $f_s$. Then we fuse other stream features to determine interactivity.

\section{VUS Dataset}
\label{sec:vus_dataset}
To the best of our knowledge, none of the existing datasets provide detailed annotations of VUS. As shown in Tab.~\ref{tab::number_dataset}, existing datasets annotate more than 1,000,000 vehicle instances. However, very few of them are in uncommon states. In order to build the VUS dataset, we drive a car to capture images in various sites (\textit{i.e.} parks, schools, hospitals, and urban roads) and in different timeframes (\textit{i.e.} morning, noon, and afternoon). We capture about 150,000 images in total. After removing the blurred and overexposed images, we finally collect 1273 vehicle instances to label.

In summary, our VUS dataset annotates \textbf{1850} VUS instances from \textbf{1441} images in real traffic sceneries and covers \textbf{10} dynamic parts (\textit{i.e.} the bonnet, the trunk, four doors, two headlights, and two taillights) and \textbf{12} uncommon states (\textit{e.g.}, turning left/right). For each vehicle instance, we manually label its 2D bounding box, instance mask, dynamic part segmentation, and state description. If there exist human interactions, we further annotate the human bounding box, the connection between the center points of the human and the vehicle, and the VHI semantics (\textit{e.g.}, getting on/out, placing/removing luggage, \textit{etc.}). We believe our VUS benchmark can effectively verify the quality of synthetic data and quantitatively evaluate the network performance.

\begin{figure}
\begin{center}
   \includegraphics[width=0.98\linewidth]{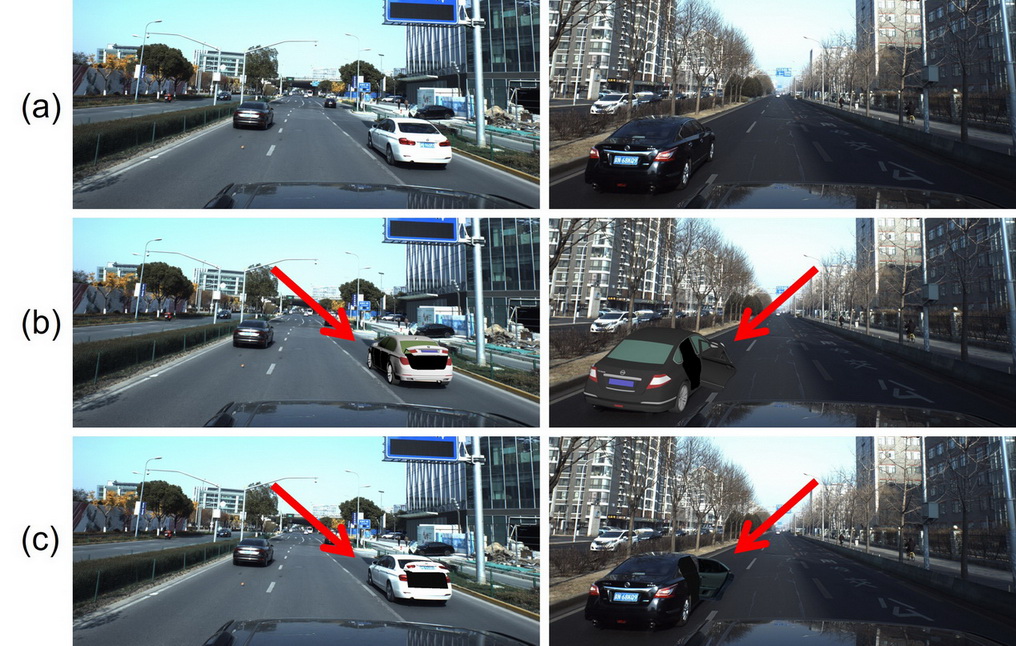}
\end{center}
   \caption{The training data of baseline methods and our method: (a) raw images; (b) rendering data; (c) synthetic data by our approach.}
\label{fig:baseline}
\end{figure}

\section{Experiments and Discussion}
\label{sec:results}
In this section, we first introduce the experimental settings and computation time. Then we describe the evaluation metric and baseline methods for comparison and present a discussion. Next, we conduct an ablation study to justify the effectiveness of our data augmentation pipeline. Furthermore, we conduct experiments to analyze the system performance, including network structure, training data number, and the rendering factors. Finally, we conduct various qualitative experiments to demonstrate the generalizability of our approach.

\subsection{Experimental Settings and Computation Time}
As described in Sec.~\ref{sec:data_augmentation}, we use the labeled 2D/3D data of ApolloCar3D to generate the visual data for the training network. To improve the diversity of the vehicles, we further annotate 100 images from other datasets (\textit{i.e.} KITTI and CityScapes). For VUS data generation, the runtime for each vehicle is about 3.0 seconds: 0.5s for 3D parts transformation and 2D projection, 0.5s for hole filling and filtering, and 2.0s for invisible region generation. For VHI data generation, the runtime is about 3.6 seconds: 3.2s for human pose adjustment and texture assignment, and 0.4s for human model rendering.

We use four commodity graphics cards (\textit{i.e.} Nvidia TITAN XPs) to train our network. The training time depends on the data number. In general, training 25K synthetic images takes 24 hours. The trained model is used for direct testing on our VUS dataset without any ``fine-turning'' strategies. As shown in Figs.~\ref{Fig:Teaser}, \ref{fig:sequence}, \ref{fig:site}, \ref{fig:city}, our network can perform fine-grained vehicle perception, which is important for safety.

\begin{table}
	\caption{Comparison with five baseline methods on 2D detection and instance segmentation on the VUS dataset. All of the numbers are the higher the better. From this table, we can see that our approach advances other baseline methods by a big margin (over 8 percentage points).}
  \begin{center}
	\begin{tabular}{c|c|c}
	\toprule[1pt]
	\textbf{Methods} & \textbf{2D Det.} (\textit{IoU}) & \textbf{Ins. Seg.} (\textit{IoU})\\    
	\hline
	\textit{1) Mask-RCNN + Existing Data} & 0.751 & 0.704 \\
	
	\textit{2) Mask-RCNN + Editing Data} & 0.758 & 0.712 \\	
	
	\textit{3) Mask-RCNN + Our Data} & 0.775 & 0.721 \\
	
	\textit{4) Our Network + Existing Data} & 0.766 & 0.713 \\
	
	\textit{5) Our Network + Rendering Data} & 0.772 & 0.719 \\
	
	\textbf{\textit{Our Network + Our Data}} & \textbf{0.862} & \textbf{0.815} \\	
	\bottomrule[1pt]
	\end{tabular}

    \label{tab::comparisons}
  \end{center}
\end{table}

\subsection{Evaluation Metric}
In Sec.~\ref{sec::baseline}, we take Mask-RCNN as a baseline network to compare with our network. Note that the proposed benchmark is only focused on VUS, while Mask-RCNN cannot distinguish between the vehicles in \textit{common/uncommon} states, which both exist in the testing data. If we use the \textit{AP} metric to evaluate this experiment, the detected common-state vehicles will decrease the precision, resulting in an inaccurate \textit{AP} value. Therefore, we compute the maximum number of \textit{IoU} values between the ground truth and the predictions to evaluate the network performance.

Different from Mask-RCNN, our network can correctly detect vehicles in uncommon states. For the ablation study in Sec.~\ref{sec::ablation_study} and the performance analysis in Sec.~\ref{sec::performance}, we choose the \textit{mAP} metric to evaluate the performance of 2D detection, instance-level segmentation, dynamic part segmentation, and VHI Detection. For state description and VHI description, we compute the match rate at each binary item between prediction state vectors and the ground truth.

\begin{table*}
    \caption{Ablation study of our data augmentation pipeline. All numbers are the higher the better. From the table, we can see that our VUS network and VHI network trained by VUS data and VHI data get the best performances on the fine-grained vehicle perception tasks.  }
  \begin{center}
	\begin{tabular}{c|c|c|c|c|c|c}
	\toprule[1pt]
	\multirow{2}{*}{\textbf{Fine-Grained Vehicle Perception}} & \multicolumn{4}{c|}{\textbf{VUS Parsing}} & \multicolumn{2}{c}{\textbf{VHI Parsing}} \\ \cline{2-7}
	& {Detection} & {Instance Seg.} & {Part Seg.} & {State Desc.} & {Detection} & {Semantics}\\ 
	\hline
	VUS Data & 0.632 & 0.485 & / & / & / & / \\
	
	VUS Data + VUS Network & 0.701 & 0.563 & 0.314 & 0.874 & / & / \\	
	
	VUS Data + VHI Data + VUS Network & 0.733 & 0.581 & 0.345 & 0.893 & / & / \\
	
	VUS Data + VHI Data + VUS Network + VHI Network & 0.733 & 0.581 & 0.345 & 0.893 & 0.678 & 0.562 \\
	\bottomrule[1pt]
	\end{tabular}

     \label{tab::ablation_study}

  \end{center}
\end{table*}

\begin{figure}
\begin{center}
   \includegraphics[width=0.97\linewidth]{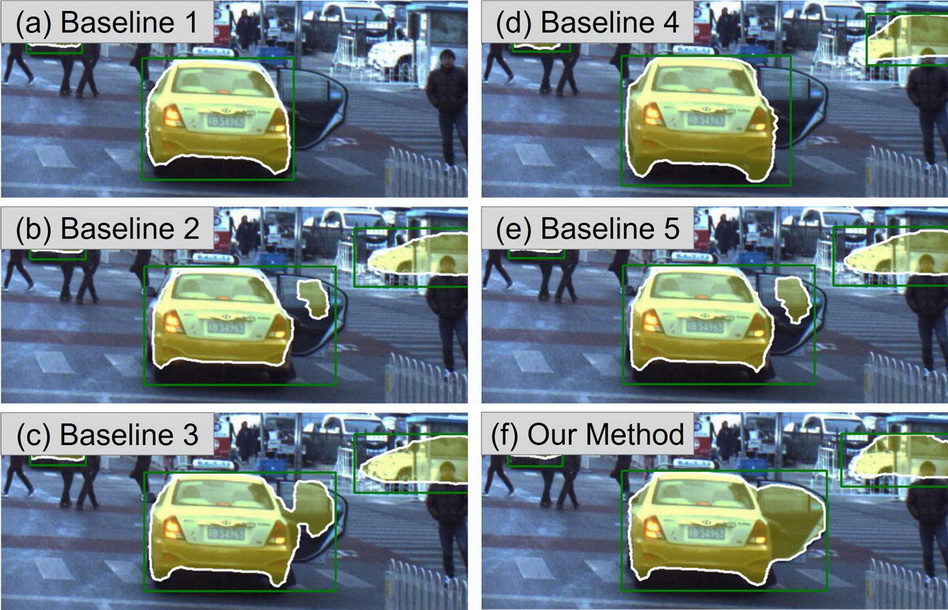}
\end{center}
   \caption{Visualization results on 2D detection and instance segmentation (five baseline methods \textit{vs.} ours).}
   \label{fig:baselinecompare}
\end{figure}

\subsection{Comparisons}
\label{sec::comparisons}
\subsubsection{Comparison on the VUS Parsing Task}
\label{sec::baseline}

Empirically, the performance of deep models largely depends on the training data and the network architecture. To verify the quality of our synthetic data, we compare it with the existing data and the rendering data. Specifically, the existing data are from the popular AD-oriented datasets, including KITTI, CityScapes, ApolloScape, and ApolloCar3D, which only provide the annotations of vehicles in common states (Fig.~\ref{fig:baseline}~(a). For the rendering data, we employ the model-based 3D rendering pipeline \cite{su2015render, prakash2018structured, tremblay2018training} to generate the training data. Following \cite{alhaijaaugmented}, we first construct 50 high-quality vehicles models with movable parts. Then we render these models with possible uncommon states to the background images. Finally, we construct the rendering dataset for the training network  (Fig.~\ref{fig:baseline}~(b).

To justify the effectiveness of our network architecture for VUS parsing, we compare it with Mask-RCNN, which is a strong baseline in 2D detection and instance segmentation. To perform a fair comparison, we set the number of training images as 25K in our experiment. In the testing phase, we directly output the results of 2D detection and instance-level segmentation on the VUS dataset.
 
The comparison results are shown in Tab.~\ref{tab::comparisons}. For better illustration, we further visualize the parsing results in Fig.~\ref{fig:baselinecompare}. The results of \textit{Baseline 1 (Mask-RCNN + Existing Data)} indicate that they can detect and segment the common-state car body but are limited to the dynamic parts. 
The results of \textit{Baseline 2 (Mask-RCNN + Rendering Data)} show that rendering data can improve the network performance compared with \textit{Baseline 1}. However, the rendering data has a natural domain gap from the real captured images. In addition, 3D rendering costs more than 10x our approach. 
The results of \textit{Baseline 3 (Mask-RCNN + Our Data)} prove that Mask-RCNN trained by our synthetic data outperforms existing datasets and rendering data. However, Fig.~\ref{fig:baselinecompare}~(c) shows that the parsing results of the visible parts are good while the reverse side of the dynamic part suffers from errors.
We then evaluate \textit{Baseline 4 (Our Network + Existing Data)} and \textit{Baseline 5 (Our Network + Rendering Data)}. The performances of both methods are slightly improved. 
Here, we emphasize that our two-backbone network is carefully designed to learn our synthetic data, especially the dynamic parts. Directly using our network cannot effectively learn other data, because they are in different domains.  Consequently, our two-backbone network trained by synthetic data gets the best performance, which advances other methods by over \textit{8\%} on both tasks. The main improvement comes from the invisible regions (Fig.~\ref{fig:baselinecompare}~(f)).

\begin{table}
    \caption{Comparison with Faster-RCNN-based HOI detection method.}

  \begin{center}
	\begin{tabular}{c|c|c}
	\toprule[1pt]
	\textbf{Tasks} & \textbf{Faster-RCNN} & \textbf{Ours} \\ 
	\hline
	VHI Detection ($mAP$) & 0.634 & \textbf{0.678} \\
	
	VHI Description ($mAP$) & 0.432 & \textbf{0.562}\\	
	\bottomrule[1pt]
	\end{tabular}
     \label{tab::VUSforVHI}

  \end{center}
\end{table}

\subsubsection{Comparison on the VHI Parsing Task}
\label{sec::hoi}
We use the VUS parsing network (Sec.~\ref{sec:vus_parsing}) as a detector to extract the input of the VHI network. Compared with the Faster-RCNN-based method, it can provide more information about the locations of the movable parts and the detailed state descriptions of VUS, which are important cues for VHI detection and detailed VHI description. In this experiment, our VHI parsing approach is compared with the Faster-RCNN-based method. We quantitatively evaluate these two methods on the VUS dataset by using equal numbers of training data. As shown in Tab.~\ref{tab::VUSforVHI}, with a movable part location and a detailed car state provided by the VUS parsing network, the VHI network achieves \textit{4.4\%} and \textit{13.0\%} improvements for interaction reasoning and detailed pattern localization, respectively.

\begin{figure}
\begin{center}
   \includegraphics[width=0.94\linewidth]{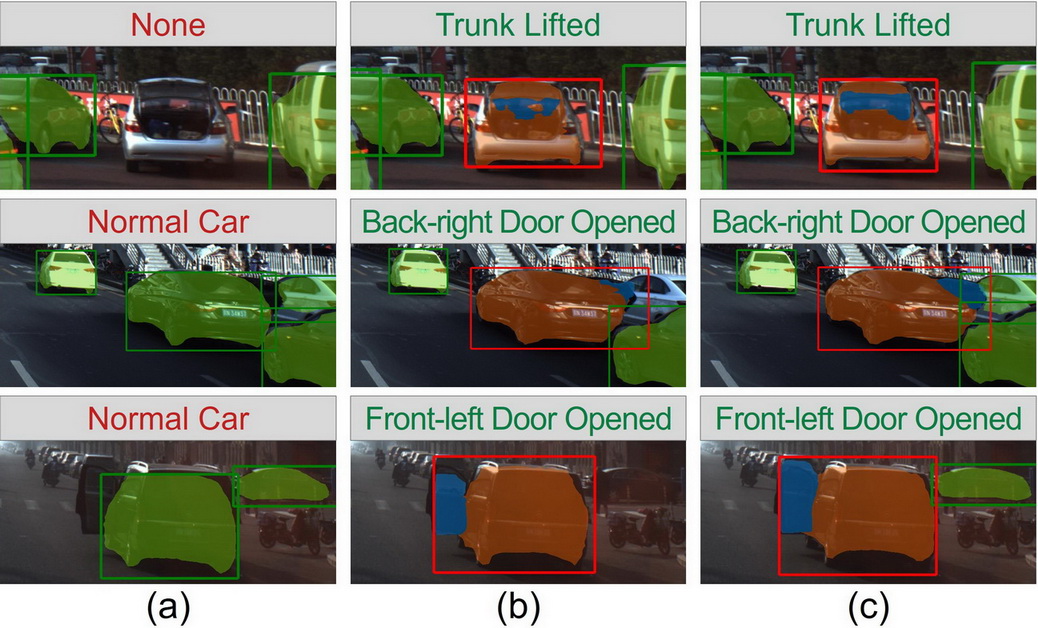}
\end{center}
   \caption{Visual results on the ablation study of our network: (a) single backbone re-trained; (b) single backbone frozen; (c) two backbones frozen. The words in red/green indicate the wrong/correct descriptions. }
\label{fig:networkresults}
\end{figure}

\subsection{Ablation Study of Our Data Augmentation Pipeline}
\label{sec::ablation_study}
Our data augmentation pipeline includes four main components: 1) VUS data generation; 2) VHI data generation; 3) multi-task network for VUS parsing; and 4) multi-stream network for VHI parsing. To justify the effectiveness of each component, we conduct an ablation study on the fine-grained vehicle perception tasks. As depicted in Tab.~\ref{tab::ablation_study}, we first use the VUS data to train a Mask-RCNN network, which is used as a baseline method (the second row). Then we train the VUS network using the VUS data (the third row). This approach not only performs the tasks of part-level segmentation and state description, but also effectively improves 2D detection and instance-level segmentation tasks by \textit{6.9\%} and \textit{7.8\%}, respectively. Next, we use the VHI data and the VUS data to train the VUS network (the fourth row). It is shown that VHI data improves the network performance (\textit{i.e.} \cite{liu20203d}) on 2D detection, instance-level segmentation, part-level segmentation, and state description by \textit{3.2\%}, \textit{1.8\%}, \textit{3.1\%}, and \textit{1.9\%}, respectively. Based on the VUS data and VHI data, we further train the VHI network to parse the vehicle-human interactions (the fifth row). The results of VHI detection and VHI semantics are \textit{67.8\%} and \textit{56.2\%}, respectively. From this ablation study, we can see that our data augmentation pipeline is robust and effective for fine-grained vehicle perception.

\subsection{Performance Analysis}
\label{sec::performance}

\subsubsection{The Impact of Our Network Structure}
The key to our network architecture is the pre-trained two backbones. Here, we conduct an ablation study to justify its effectiveness. As shown in Tab.~\ref{tab::backbone}, we first re-train the single backbone, which is a common strategy in most deep networks (\textit{e.g.}, \cite{he2017mask}). The results show that it is barely able to predict the correct class of VUS, leading to bad performance on these tasks (Fig.~\ref{fig:networkresults}~(a)). We then freeze the single backbone pre-trained on COCO during the synthetic data training. The performance is improved because we have relieved the over-fitting problem. However, the frozen backbone cannot extract adequate features (Fig.~\ref{fig:networkresults}~(b)).
In contrast, our two backbones, which are pre-trained on a car detection task and a general task can not only extract adequate features but also avoid the over-fitting problem. It achieves the best performance (Fig.~\ref{fig:networkresults}~(c)).

\begin{figure}
\begin{center}
   \includegraphics[width=0.91\linewidth]{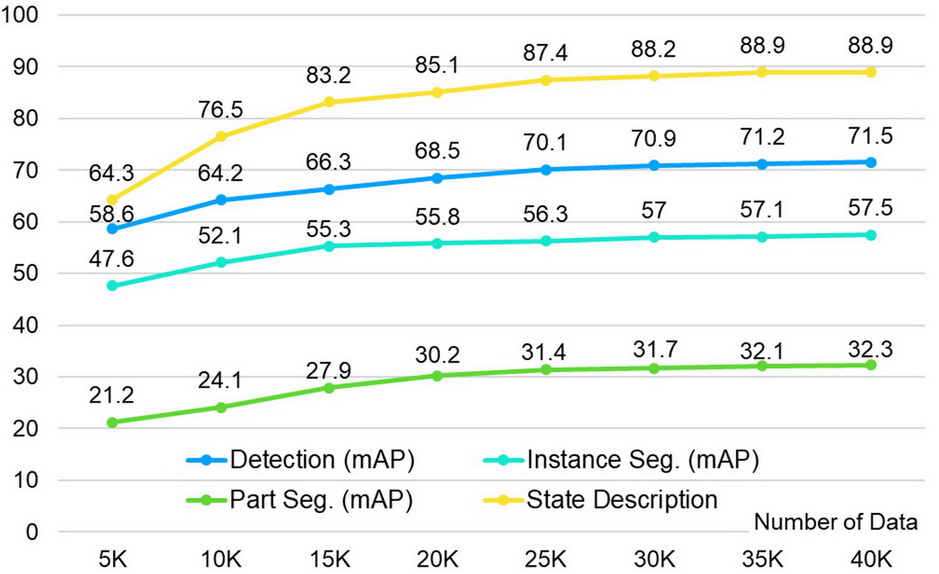}
\end{center}
   \caption{The performance of our two-backbone network with different numbers of training data.}
\label{fig:trainingnumber}
\end{figure}

\begin{table}
    \caption{Ablation study of our network on 2D detection, instance segmentation, part segmentation, and state description. ``S. B. Retrained'' indicates single backbone retained, ``S. B. Frozen'' means single backbone frozen, and ``T. B. Frozen'' indicates two backbones frozen. We see that the strategy of two backbones frozen gets the best performance.}
  \begin{center}
	\begin{tabular}{c|c|c|c}
	\toprule[1pt]
	
	\textbf{Tasks} & \textbf{S. B. Retrained} & \textbf{S. B. Frozen} & \textbf{T. B. Frozen}  \\ 
	\hline
	Detection ($mAP$) & 0.136 & 0.672 &  \textbf{0.701} \\
	
	Ins. Seg. ($mAP$) & 0.114 & 0.516 & \textbf{0.563}\\	
	
	Part Seg.($mAP$) & 0.144 & 0.273 & \textbf{0.314}\\
	
	State Description & 0.149 & 0.837 & \textbf{0.874}\\
	
	\bottomrule[1pt]
	\end{tabular}

	 \label{tab::backbone}

  \end{center}
\end{table}

\subsubsection{The Impact of the Number of Training Data}
Empirically, the performance of deep networks largely relies on the number of training data. To study the relationship between them, we train the network with different numbers of data from 5K to 40K with an interval of 5K. As shown in Fig.~\ref{fig:trainingnumber}, from 5K to 25K, the network performance is significantly improved. From 25K to 40K, however, the network is not sensitive to the number of training data. 
There are two main reasons: 1) network capability (we cannot always improve the network performance by adding to the number of training data) and 2) test data (there exist some features or cases in the test data that are not well learned by the network). In practice, we set the number of training data to \textit{25K}, which is a good compromise of efficiency and accuracy.

\begin{figure}
\begin{center}
   \includegraphics[width=1.0\linewidth]{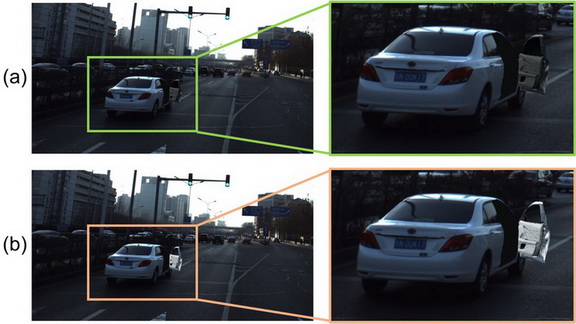}
\end{center}
   \caption{The synthetic VUS results with (a) and without (b) pre-calculated environment map.}
\label{fig:envmap}
\end{figure}

 \begin{table}
    \caption{The impact of the environment map. The table shows that the environment map significantly improves the network performance. In particular, the dynamic part segmentation gets a 9.3 percent improvement.}

  \begin{center}
	\begin{tabular}{c|c|c}
	\toprule[1pt]
	\textbf{Tasks} & \textbf{w/o Env. Map} & \textbf{with Env. Map}  \\ 
	\hline
	2D Detection (\textit{mAP}) & 0.688 & \textbf{0.701} \\
	
	Ins. Seg. (\textit{mAP}) & 0.538 & \textbf{0.563}\\	
	
	Part Seg. (\textit{mAP}) & 0.221 & \textbf{0.314}\\
	
	State Description & 0.844 & \textbf{0.874}\\
	\bottomrule[1pt]	
	\end{tabular}
    \label{tab::map}

  \end{center}
\end{table}

 \begin{figure*}
\begin{center}
   \includegraphics[width=1.0\linewidth]{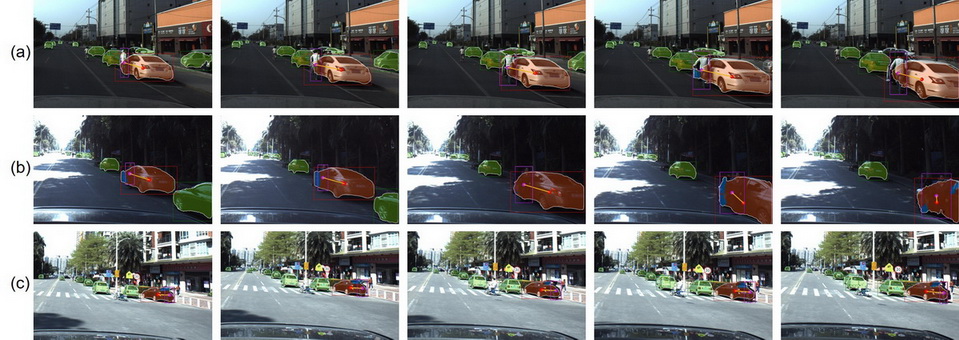}
\end{center}
   \caption{Fine-grained parsing results of VUS and VHI using our approach in three video sequences of street-view images.}
\label{fig:sequence}
\end{figure*}

\begin{figure}
\begin{center}
   \includegraphics[width=0.85\linewidth]{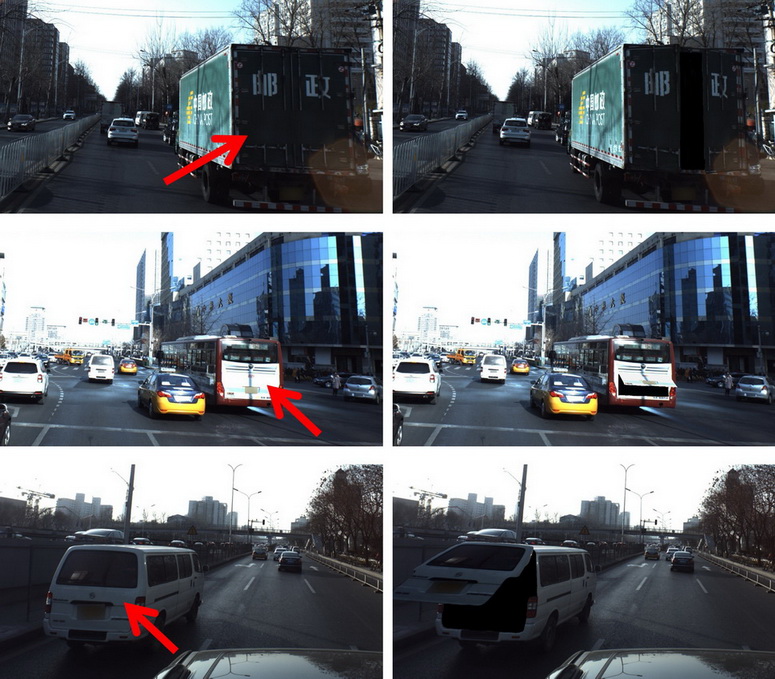}
\end{center}
   \caption{More synthetic results of vehicles using our approach. The vehicles' types include truck, bus, and van.}
\label{fig:more_editing}
\end{figure}

\begin{figure*}
\begin{center}
   \includegraphics[width=1.0\linewidth]{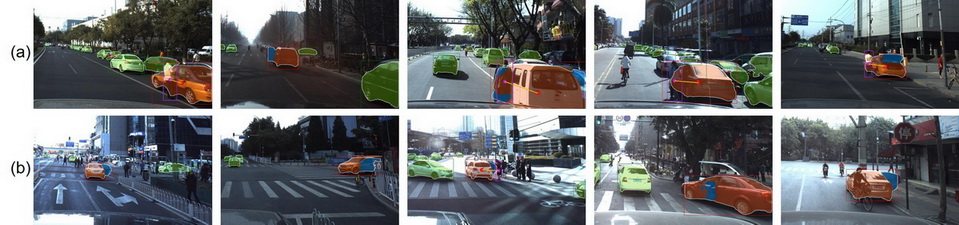}
\end{center}
   \caption{Fine-grained parsing results of VUS and VHI using our method in different sites: (a) the side of the road; (b) the crossroads. }
\label{fig:site}
\end{figure*}

\begin{figure*}
\begin{center}
   \includegraphics[width=1.0\linewidth]{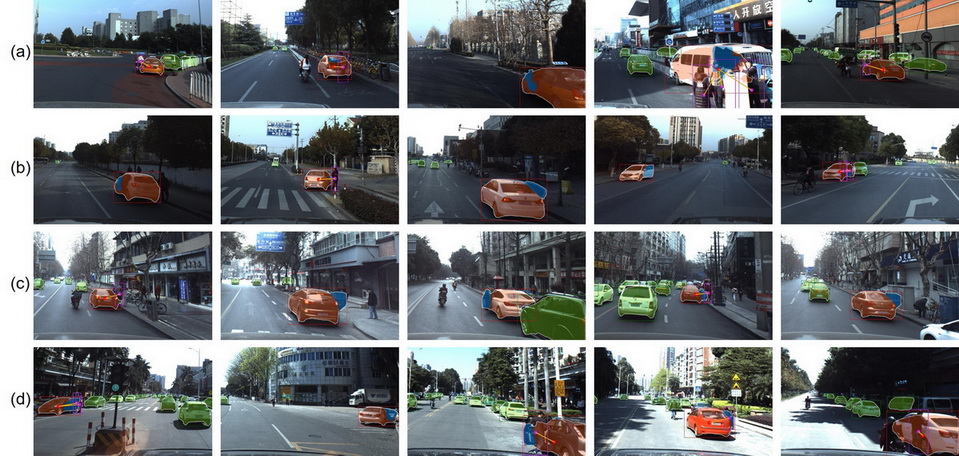}
\end{center}
   \caption{Fine-grained parsing results of VUS and VHI using our method in different cities: (a) Beijing; (b) Shanghai; (c) Chengdu; (d) Guangzhou.}
\label{fig:city}
\end{figure*}

\subsubsection{The Impact of the Environment Map}
As described in Sec.~\ref{sec:data_augmentation_VUS}, the invisible region data are generated by 3D part rendering using the environment map. In the real traffic scenarios, the environment map (or illumination) plays an important role in determining whether the rendered region is compatible with the surroundings. Here, we utilize the same number of reverse side data with/without an environment map (shown in Fig.~\ref{fig:envmap}) to train our network and evaluate it on the proposed VUS dataset. As shown in Tab.~\ref{tab::map}, the data rendered by the environment map significantly improves the network performance. In particular, the dynamic part segmentation is improved by \textit{9.3\%}.

\subsection{Qualitative Evaluation}
It is essential to conduct quantitative evaluations and comparisons, as in Sec.~\ref{sec::comparisons}, Sec.~\ref{sec::ablation_study}, and Sec.~\ref{sec::performance}. However, for AD, robustness and generalizability are the most critical issues. In this section, we conduct various qualitative experiments to demonstrate the effectiveness of our approach.

\subsubsection{Generalizability of Our Approach}
Our data augmentation approach is generic and can be transferred to other objects such as the human body and furniture. These objects are constructed out of parts, and each part has a corresponding motion axis. Based on the 2D-3D alignment datasets, ApolloCar3D (vehicle), Unite the People (human body), and Pixel3D (furniture), we can generate diverse results using our approach. In this paper, we focus on vehicle augmentation and perception. Fig.~\ref{fig:more_editing} shows the synthetic results of different vehicle types, including truck, bus, and van.

\subsubsection{Fine-grained Parsing on Video Sequence}
Fig.~\ref{fig:sequence} shows the fine-grained perception results produced by our approach in three different sequences of street-view images. The common vehicles are shown in green, which can be obtained by conventional methods of detection and segmentation. In contrast, our approach can perceive the VUS cases (in red bounding boxes) as well as the VHI cases (in pink bounding boxes). Both are important for the safety of the autonomous vehicle. It is clear that accurate and robust instance segmentation results can be estimated by our network. Note that these images are captured by a moving car, which is different from our training data. The results are generated without network ``re-training'' or ``fine-tuning'', which justifies the advantages of our approach.

\subsubsection{Parsing Results on Different Sites}
In Fig.~\ref{fig:site}, we demonstrate the fine-grained vehicle perception results on different sites, particularly the side of the road and the crossroads. Usually, vehicles are not allowed to stop at these sites with doors/trunks open for getting in/out or placing/removing luggage. However, we still find a number of dangerous behaviors from drivers and pedestrians. Our approach can perceive these VUS cases and VHI cases in advance, which is very important for the planning and decision modules of AD systems.

\subsubsection{Parsing Results on Different Cities}
To demonstrate the robustness and generalizability of our approach, we conduct experiments in various cities, which have different weather conditions, different buildings, different roads, and different plants. Fig.~\ref{fig:city} shows the fine-grained vehicle perception results for different cities, including Beijing, Shanghai, Chengdu, and Guangzhou. Note that these source images are different from the training data and our network is not ``re-trained'' or ``fine-tuned'' to fit the test data. From this experiment, we find that VUS cases and VHI cases vary wildly in traffic scenarios. Our approach fills the missing piece by providing detailed vehicle parsing and ``vehicle-human'' interaction semantics.

\section{Conclusion and Limitation}
In this paper, we make the first attempt to analyze vehicles in uncommon states (VUS). Instead of annotating a large number of images, we present a visual data augmentation approach that takes advantage of 3D parts. Our data augmentation approach is light-weight but high-efficiency, which can automatically generate a large number of training data. To perform fine-grained vehicle perception, we present a multi-task network for VUS parsing and a multi-stream network for VHI parsing. For benchmarking, we build the first VUS dataset containing 1441 real images (1850 car instances) with fine-grained annotation. The experimental results show that our synthetic data and the proposed deep networks perform well on VUS.

Nevertheless, there are a number of limitations, which we will tackle in future work. First, the uncommon states analyzed in this paper are limited to vehicles. Some other objects, such as buildings and roads, require more attention. Second, the outputs of our network are mostly 2D results. We will extend this work to the 3D space, using 3D detection, 3D localization, and 3D reconstruction. Finally, we will research VUS and VHI on the video sequences and multiple sensors fusion data (\textit{e.g.}, cameras, Lidar, and Radar).

\appendices


%
%

\ifCLASSOPTIONcaptionsoff
  \newpage
\fi



\bibliographystyle{IEEEtran}
\bibliography{bibliography}

\begin{IEEEbiography}[{\includegraphics[width=1in,height=1.25in,clip,keepaspectratio]{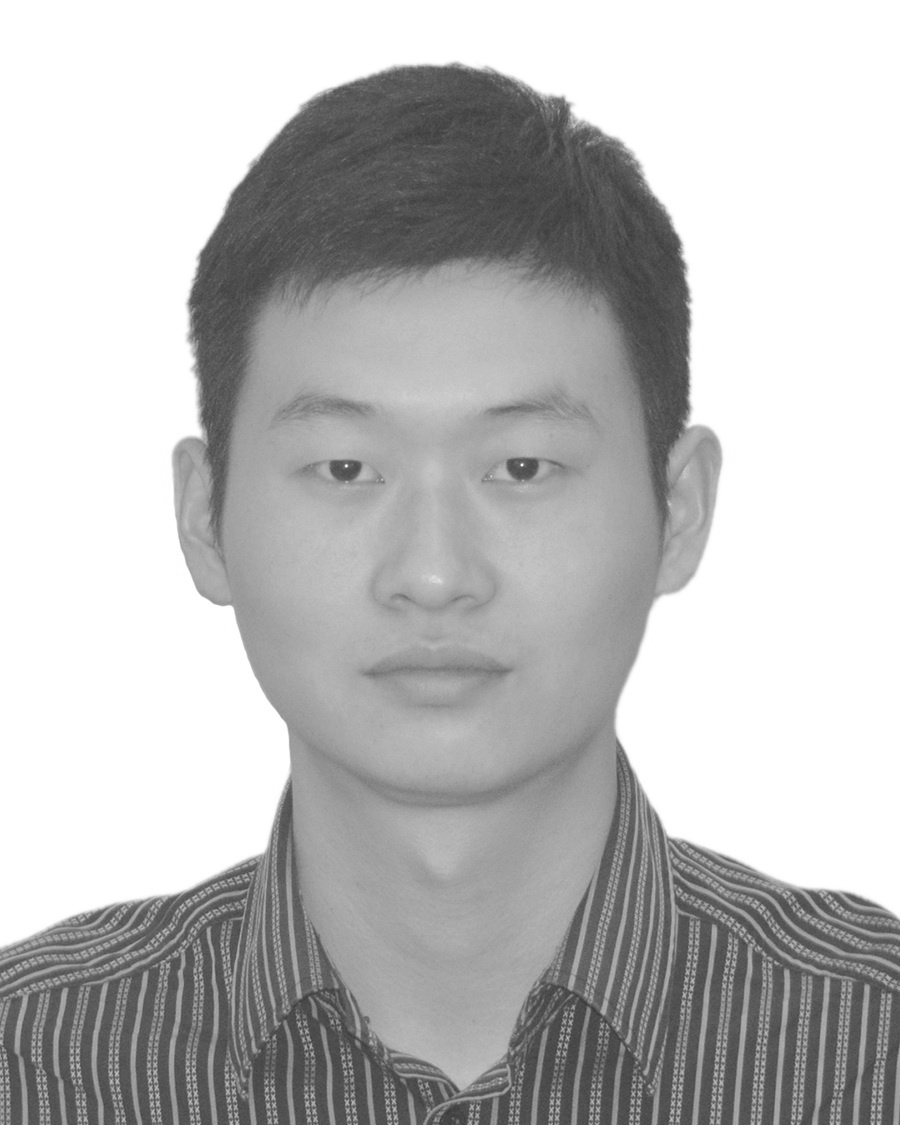}}]{Feixiang Lu}
is currently a senior researcher at the Robotics and Autonomous Driving Laboratory (RAL), Baidu Research, Baidu Inc., China. He obtained his Ph.D. degree in Computer Science from Beihang University in 2019. He received the B.S. degree from Beijing University of Posts and Telecommunications in 2013. His research interests include 3D reconstruction, scene parsing, and their applications in autonomous driving. He received the Best Paper Award at the Computer Graphics International (CGI) in 2018. 
\end{IEEEbiography}

\begin{IEEEbiography}[{\includegraphics[width=1in,height=1.25in,clip,keepaspectratio]{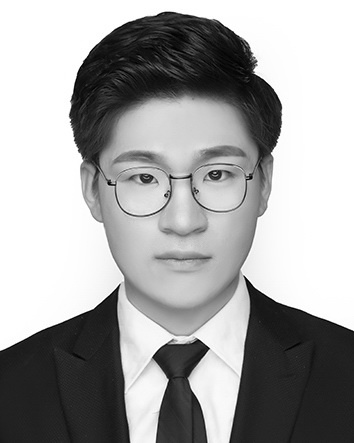}}]{Zongdai Liu}
is currently a research intern in Robotics and Autonomous Driving Lab, Baidu Research, Baidu Inc., China. He is also a master student in Beihang University. He received the B.S. degree from Taiyuan University of Technology. His research interests include 3D detection, scene parsing, and their applications in autonomous driving.
\end{IEEEbiography}

\begin{IEEEbiography}[{\includegraphics[width=1in,height=1.25in,clip,keepaspectratio]{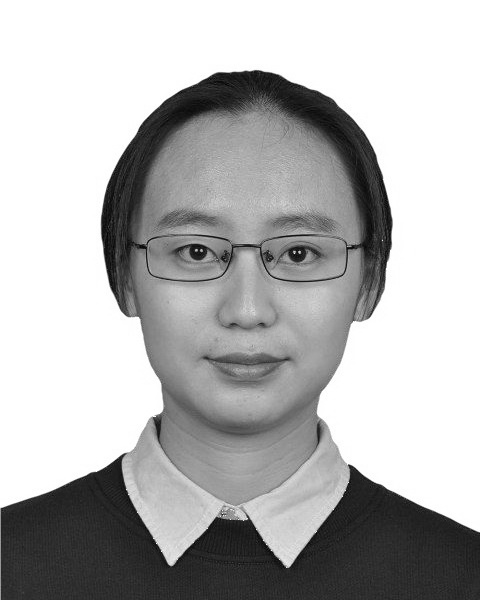}}]{Hui Miao}
is currently a master student in School of Computer Science and Engineering, Beihang University. She received the B.S. degree from Beijing Information Science \& Technology University. And her research interests include 3D detection, data augmentation and their applications in autonomous driving.
\end{IEEEbiography}

\begin{IEEEbiography}[{\includegraphics[width=1in,height=1.25in,clip,keepaspectratio]{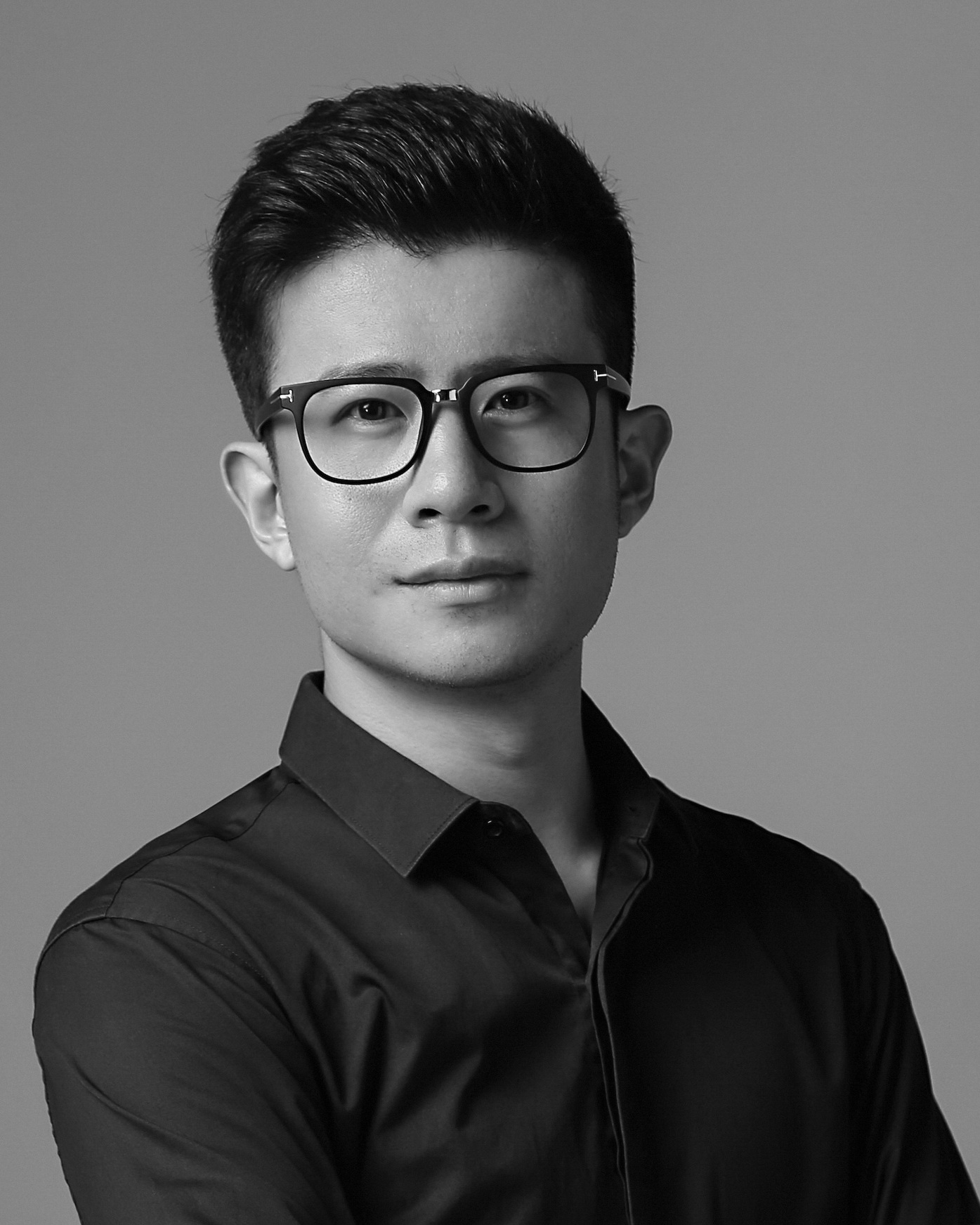}}]{Peng Wang}
is currently a senior research scientist at Bytedance. He graduated from UCLA advised by Professor Alan Yuille. Before UCLA, he obtained his B.S and M.S degree on machine intelligence from Peking University. 
His research interest is especially on perception of autonomous vehicle/moving robot, such as learning depth and segmentation from stereo, monocular videos/images, fusing multiple sensors such as video and LiDAR. He has published over 20 papers on major computer vision and machine learning conferences such as CVPR, ICCV, ECCV and NIPS.
\end{IEEEbiography}

\begin{IEEEbiography}
[{\includegraphics[width=1in,height=1.25in,clip,keepaspectratio]{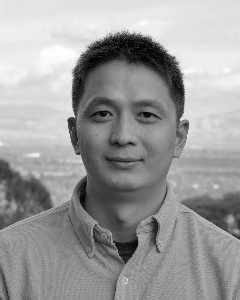}}]{Liangjun Zhang}
is the lead of Robotics and Autonomous Driving Lab (RAL) of Baidu Research. Dr. Zhang received his PhD in computer science from the University of North Carolina at Chapel Hill and MS/BS from Zhejiang University. He was a NSF computing innovation fellow in the computer science department at Stanford University. His research interests span robotics, autonomous driving, simulation, geometric computing and computer graphics. He has received a number of awards including the Best Paper Award at the International CAD Conference and the UNC Linda Dykstra Distinguished PhD Dissertation Award. 
\end{IEEEbiography}

\begin{IEEEbiography}
[{\includegraphics[width=1in,height=1.25in,clip,keepaspectratio]{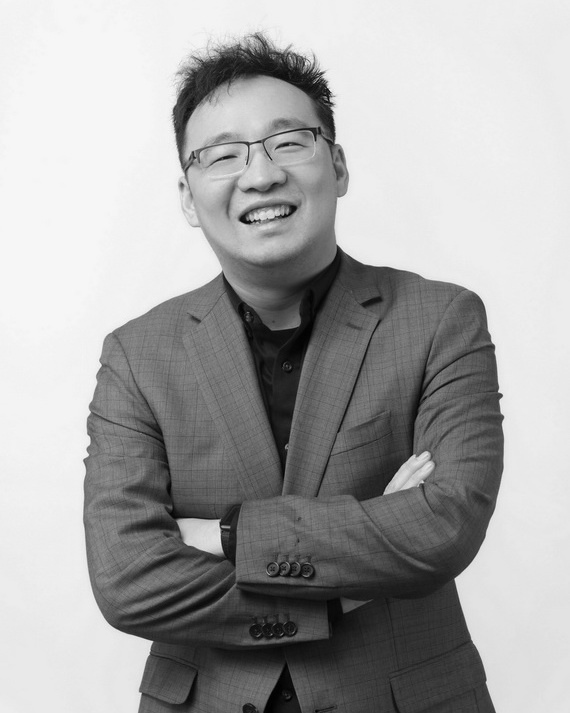}}]{Ruigang Yang} is the CTO of Inceptio. He is a renowned computer vision scientist. He has published over 100 papers with an H-index of 58. He was an Associate Editor for IEEE T-PAMI. He has served as Area Chairs for premium vision conferences (such as ICCV/CVPR), and serve as a Program Chair for CVPR 2021. He has made a number of world-class scientific research achievements in the field of 3D reconstruction and 3D data analysis. Before joining Inceptio, he served as the chief scientist for 3D computer vision and led the Robot and Autonomous Driving Laboratory of Baidu Research. Before that, he was a tenured full professor at the University of Kentucky, USA. He holds a PhD in Computer Science from University of North Carolina at Chapel Hill (UNC-CH), a MS in Computer Science from Columbia University, and finished his undergraduate study from Tsinghua University.

\end{IEEEbiography}

\begin{IEEEbiography}
[{\includegraphics[width=1in,height=1.25in,clip,keepaspectratio]{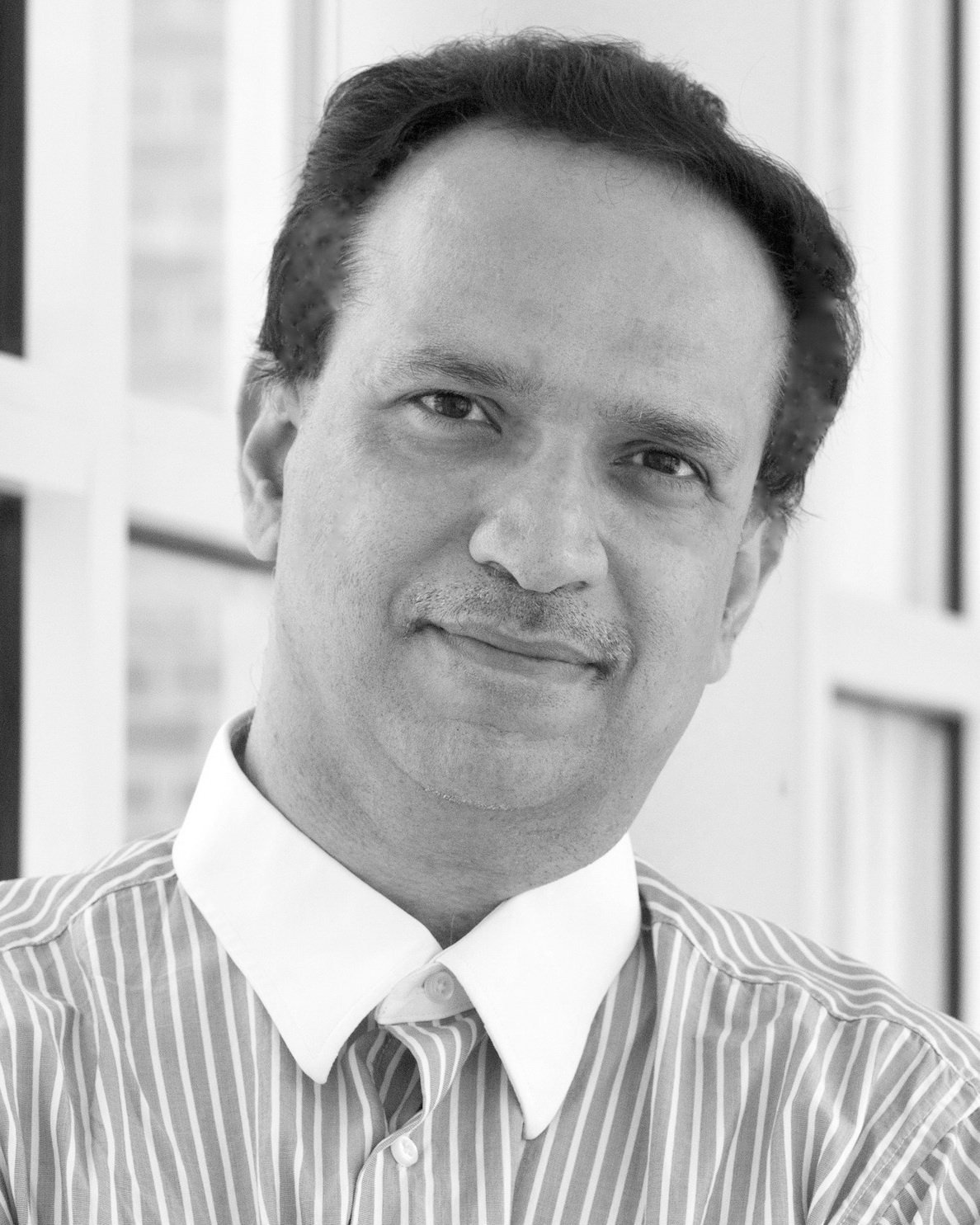}}]{Dinesh Manocha} 
is currently the Paul Chrisman Iribe Chair of Computer Science and Electrical \& Computer Engineering at the University of Maryland at College Park. Earlier he was the Matthew Mason/Phi Delta Theta Distinguished Professor of Computer Science at the University of North Carolina at Chapel Hill. He co-leads a major research group with more than 20 members on geometric and simulation algorithms with applications to computer graphics, robotics and virtual environments. He has published more than 500 papers in the leading conferences and journals in computer graphics, robotics, computational geometry, databases, multimedia, high performance computing and symbolic computation, and received 16 best paper and time of test awards. He has also served as program committee member or program chair of more than 120 leading conferences in these areas. Moreover, he has given more than 110 invited or keynote talks at conferences and distinguished lectures at other institutions. Manocha has served as a member of the editorial board or guest editor of eleven leading journals in computer graphics, robotics, geometry processing and scientific computing. He is a co-inventor of 9 patents, several of which have been licensed to industry. He has won many awards including NSF Career Award, ONR Young Investigator Award, Sloan Fellowship, IBM Fellowship, SIGMOD IndySort Winner, Honda Research Award, UNC Hettleman Prize, etc. He is a Fellow of ACM, AAAS, AAAI, and IEEE, and received the Distinguished Alumni Award from Indian Institute of Technology, Delhi.
\end{IEEEbiography}

\begin{IEEEbiography}
[{\includegraphics[width=1in,height=1.25in,clip,keepaspectratio]{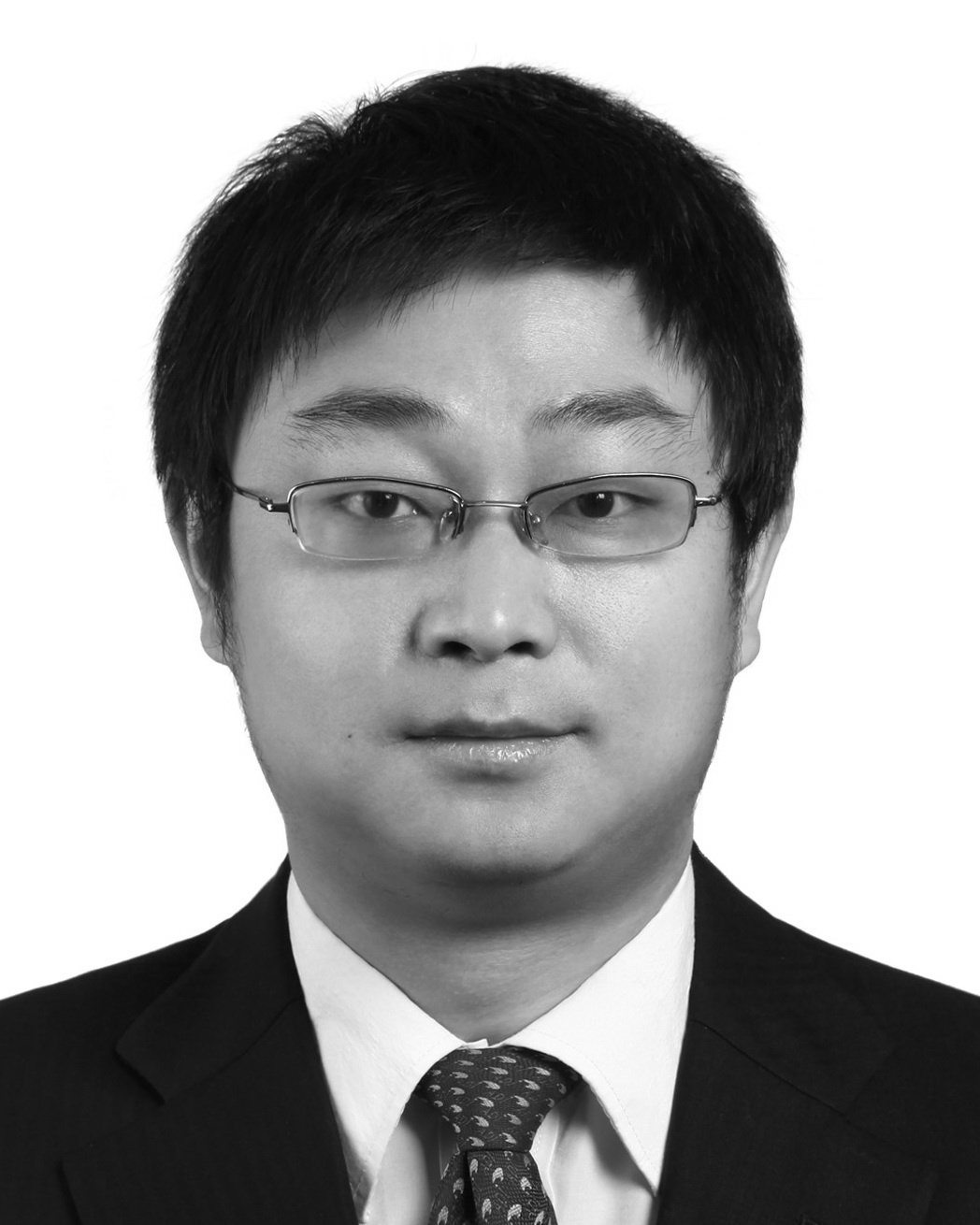}}]{Bin Zhou} 
is currently an Associate Professor with the State Key Laboratory of Virtual Reality Technology and Systems, School of Computer Science and Engineering, Beihang University. He is also an Associate Professor with Peng Cheng Laboratory, Shenzhen, China. He received his B.S. and Ph.D. degrees in Computer Science from the Beihang University, China, in 2006 and 2014, respectively. His research interests include Computer Graphics, Virtual Reality, Computer Vision and Robotics. He is the editorial board member of Virtual Reality \& Intelligent Hardware. He served on the program committee of multiple conferences and workshops including ISMAR 2019, AsiaHaptics 2020, VRST 2015, ICVRV 2017$\sim$2020, ChinaVR 2017$\sim$2020. He has received a number of awards including the Best Paper Award at Computer Graphics International 2018. 
\end{IEEEbiography}

\end{document}